\documentclass{article}

\usepackage{arxiv}

\usepackage{microtype}
\usepackage{graphicx}
\usepackage{subfigure}
\usepackage{booktabs}

\usepackage{algorithm}
\usepackage{algorithmic}
\usepackage{times}
\usepackage{soul}
\usepackage{url}
\usepackage[utf8]{inputenc}
\usepackage{amsmath}
\usepackage{amsthm}
\usepackage{amsfonts}
\newtheorem{theorem}{Theorem}
\usepackage{graphicx,subfigure,caption}
\usepackage{caption}
\usepackage{wrapfig}
\usepackage{comment}
\usepackage{enumitem}
\usepackage{xcolor}

\usepackage{hyperref}
\usepackage{xr}

\title{Associative Memories via Predictive Coding}

\author{
Tommaso Salvatori\thanks{Equal Contribution}
\\
Department of Computer Science\\
University of Oxford, UK\\
\texttt{tommaso.salvatori@cs.ox.ac.uk} \\
\And
Yuhang Song\footnotemark[1]
\\
Department of Computer Science\\
University of Oxford, UK\\
\texttt{yuhang.song@some.ox.ac.uk} \\
\And
Yujian Hong\footnotemark[1]\\
Department of Computer Science\\
University of Oxford, Oxford, UK \\
\texttt{yujian.hong@cs.ox.ac.uk}
\And
Simon Frieder\\
Department of Computer Science\\
University of Oxford, Oxford, UK \\
\texttt{frieder.simon@cs.ox.ac.uk}
\And
Lei Sha\\
Department of Computer Science\\
University of Oxford, Oxford, UK \\
\texttt{lei.sha@cs.ox.ac.uk}
\And
Zhenghua Xu
\\
State Key Laboratory \\
Hebei University of Technology, Tianjin, China \\
\texttt{zhenghua.xu@hebut.edu.cn} \\

\And
Rafal Bogacz \\
MRC Brain Network Dynamics Unit \\
University of Oxford, UK \\
\texttt{rafal.bogacz@ndcn.ox.ac.uk}\\
\And
Thomas Lukasiewicz \\
Department of Computer Science\\
University of Oxford, UK\\
\texttt{thomas.lukasiewicz@cs.ox.ac.uk}}

\begin{document}
\maketitle

\begin{abstract}\vspace*{-0.5ex}
Associative memories in the brain receive and store patterns of activity registered by the sensory neurons, and are able to retrieve them when necessary. Due to their importance in human intelligence, computational models of associative memories have been developed for several decades now. They include autoassociative memories, which allow for storing data points and retrieving a stored data point $s$ when provided with a noisy or 
partial variant of $s$, and heteroassociative memories, able to store and recall multi-modal data.
%
%
%
%
In this paper, we present a novel neural model for realizing associative memories, based on a hierarchical generative network that receives external stimuli via sensory neurons. This model is trained using predictive coding, an error-based learning algorithm inspired by information processing in the cortex. 
To test the capabilities of this model, we perform multiple retrieval experiments from both corrupted and incomplete data points. In an extensive comparison, we show that this new model outperforms in retrieval accuracy and robustness popular associative memory models, such as  autoencoders trained via backpropagation, and modern Hopfield networks. In particular, in completing partial data points, our model achieves remarkable results on natural image datasets, such as ImageNet, with a surprisingly high accuracy, even when only a tiny fraction of pixels of the original images is presented. Furthermore, we show that this method is able to handle multi-modal data, retrieving images from descriptions, and vice versa. We conclude by discussing the possible impact of this work in the neuroscience community, by showing that our model provides a plausible framework to study learning and retrieval of memories in the brain, as it closely mimics the behavior of the hippocampus as a memory index and generative model.
\vspace*{-0.5ex}\end{abstract}

\section{{Introduction}}

Through our life, we learn a huge number of associations between concepts: the taste of a particular food, the meaning of a gesture, or to stop when we see a red light. Every time we acquire new information of this kind, it gets stored in our long-term memory, situated in distributed networks of brain areas \cite{Squire91}. In particular, visual memories are stored in a hierarchical network of visual and associative areas \cite{Felleman91}. These regions learn progressively more abstract representation of visual stimuli, so they participate in both perception and memory as each area memorizes relationships present in their inputs \cite{murray07}. Accordingly, early visual areas learn common regularities present in the stimuli \cite{rao1999predictive}, while at the top of this hierarchy, associative areas (such as hippocampus, entorhinal cortex, and perirhinal cortex) store the relationships between extracted features, which encode an entire stimulus or episode \cite{Mcclelland95}. The memory system of the brain is able to both recall complex memories \cite{Squire91, Barron20}, and use them to generate predictions to guide behavior \cite{stachenfeld17}. Learning in these associative memories shapes our understanding of the world around us and builds the foundations of human intelligence.

Building models that are able to store and retrieve information has been an important direction of research in artificial intelligence.
Particularly, such models include (auto)associative memories (AMs), which allow for the storage of data points and their {contents-based retrieval}, i.e., for retrieving a stored data point $s$  from a corrupted or a partial variant of $s$. One way to realize AMs is to store data points as attractors, so that they can be easily recovered via an energy minimization process when presenting their corrupted variants~\cite{Hopfield82,krotov2019unsupervised}. Classic AMs include Hopfield networks and their modern formulation, called modern Hopfield networks (MHNs) \cite{Krotov16}. The latter are one-shot learners, which are able to store exponentially many memories, and to perfectly retrieve them. However, the retrieval process often fails when dealing with complex data, such as natural images. Recent works have shown that overparametrized autoencoders (AEs) are excellent AMs as well. Particularly, when training an AE to generate a specific point $s$ when $s$ itself is presented as an input, it gets stored as an attractor~\cite{radhakrishnan19}.

In this work, we present a novel AM model that is based on a new energy-based  generative approach. This AM model differs from standard Hopfield networks, as it is trained using predictive coding (PC), which is a biologically plausible learning algorithm inspired by learning in the visual cortex~\cite{rao1999predictive}. The idea that PC may naturally be related to AMs is inspired by recent works showing that the generative neural architecture that connects the hippocampus to the neocortex is based on an error-driven learning algorithm, which can be interpreted with a PC framework \cite{Barron20,Liu18}. From a machine learning perspective, predictive coding networks (PCNs) are able to perform both supervised and unsupervised tasks with a high accuracy \cite{rao1999predictive, whittington2017approximation}, and are completely equivalent to backpropagation when trained with a specific algorithm~\cite{Song2020,Salvatori2021,salvatori2021any}.

We show that the new AM model is not only interesting from a neuroscience perspective, but it also outperforms popular AM models when it comes to the retrieval of complex data points. Our results can be briefly summarized as follows:
\begin{itemize}
    \item We define generative PCNs and empirically show that they store training data points as attractors of their dynamics by demonstrating that they can restore original data points from corrupted versions. In an extensive comparison of the new AM model against standard AEs, the new model considerably outperforms AEs 
   (in~storage capacity, retrieval~accuracy, and robustness) when tested on neural networks of the same size.
    

    \item The reconstruction of incomplete data points is a challenging task in the field of AMs. Our model naturally solves the task of reconstructing complex and colored images with a surprisingly high accuracy, outperforming autoencoders and MHNs by a large margin on Tiny ImageNet and CIFAR10. We also test our model on ImageNet, perfectly reconstructing single pictures even after removing all but $1/{8}$ of the original image. We then show that to increase the overall capacity of the model, it suffices to add additional layers. 
    
    \item We show that our model is also able to handle multi-modal data, and hence perform hetero-associative memory experiments. Particularly, we train a model to memorize captioned images, where the captions are taken from a dictionary of $1000$ words, and use the description to retrieve the original image and vice versa. Note that retrieving the images from the captions implies retrieving $3072$ pixels, using a vector of only $25$ dimensions, which corresponds to less than $1\%$ of the total information. We show that other AM models fail in performing this complex task.
\end{itemize}

\section{Generative Predictive Coding Networks} 

We now briefly recall  {predictive coding networks} (PCNs), and we introduce generative PCNs, which are the underlying neural model for the novel AMs introduced in the subsequent section.



\begin{figure}[t]
\centering
    \includegraphics[width=0.65\textwidth]{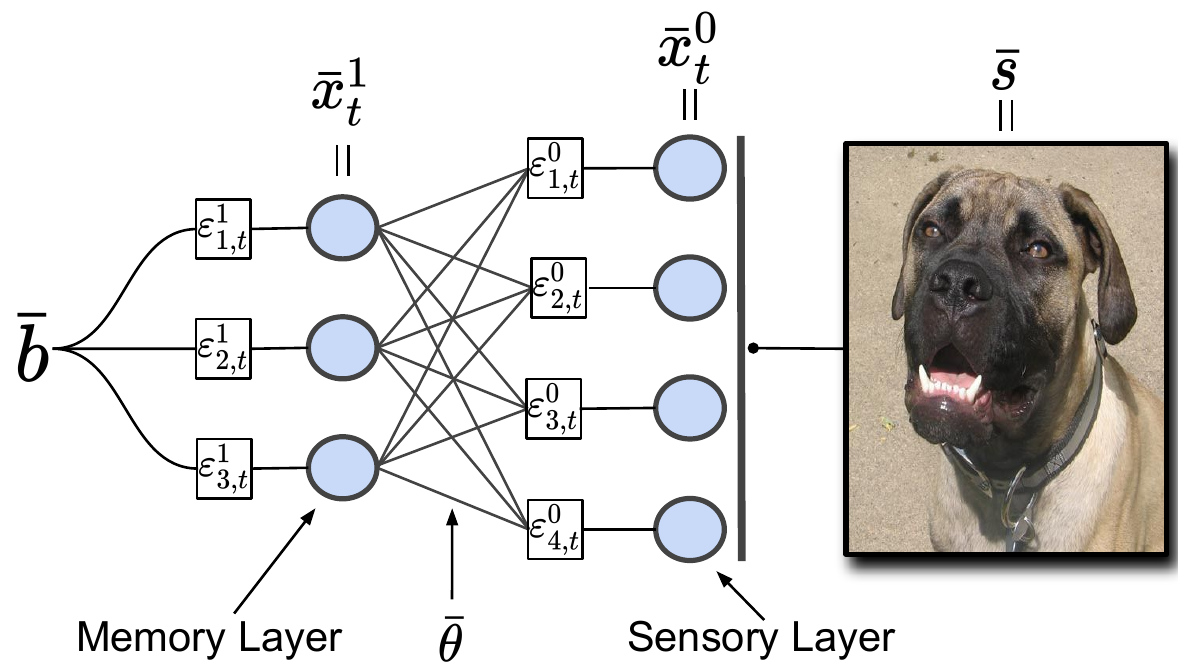}
\vspace*{-1ex}
\caption[short]{ Generative PCN with $2$ layers. Each line linking neurons denotes a pair of connections (excitatory in left direction, and inhibitory in right direction). At $t=0$, the value nodes of the output layer are fixed to the vector representing the pixels of the figure. Then, during the learning phase, both weight parameters and value nodes are updated.  }
\label{fig:eff}\vspace*{-1ex}
\end{figure}

\begin{algorithm}[t]
    \caption{Learning to generate $\bar s$ with IL} \label{algo:IL}
    \begin{algorithmic}[1]
    \REQUIRE  $\overline{x}_{0}$ is fixed to $\overline{s}$.
    \FOR{$t=0$ to $T$}
        \FOR{each neuron $i$ in each level $l$}
            \STATE Update $x^{\scriptscriptstyle {l}}_{i,t}$ to minimize $E_{t}$.
        \ENDFOR
    \ENDFOR
    \STATE Update each  $\theta^{\scriptscriptstyle {l+1}}_{i,j}$ and $b_i$ to minimize $E_{t}$  \ \  
                {
                }
    \end{algorithmic}
\end{algorithm}

Deep neural networks have a multi-layer structure, where each layer is formed by a vector of 
neurons~\cite{rumelhart1986learning}. While in standard deep learning the goal is to minimize the error on a specific layer, PC defines an error in every layer of the network, minimized by gradient descent on a global energy function \cite{rao1999predictive}.
Particularly, let $M$ be a PCN with $L-1$ fully connected layers of dimension $n$, followed by a fully connected layer of dimension $d$. We call the $d$-dimensional layer \emph{sensory} layer (indexed as layer~$0$), which biologically corresponds to sensory neurons (see Fig \ref{fig:eff}). We call the most internal layer (layer $L$) \emph{memory}, which is equipped with an $n$-dimensional memory vector $b$. Every layer $l$ contains value nodes $x^{\scriptscriptstyle {l}}_{i,t}$,~and every pair of consecutive layers is connected via weight 
matrices~$\bar \theta^l$, which represent the synaptic weights between neurons of different layers. The value nodes, the weight matrices, and the memory vector are all trainable parameters of the model.
The signal passed from layer $l+1$ to layer~$l$, called prediction $\bar \mu^l_t$,  is computed as follows:\vspace*{-2ex}
%

{\small\begin{equation}
\,\mu^{\scriptscriptstyle {l}}_{i,t}
= \begin{cases}
{\textstyle\sum}_{j=1}^{n^{\scriptscriptstyle {l+1}}} \theta^{\scriptscriptstyle {l+1}}_{i,j} f ( x^{\scriptscriptstyle {l+1}}_{j,t}) & \!\!\mbox{if } 0 \,\leq\ l \,{<}\ L\\

b & \!\!\mbox{if } l = L\,,
\end{cases}
\label{eq:pcn-forward-varepsilon}
\end{equation}}%
%
\noindent where $f$ is a non-linear activation function. To conclude, the difference between the value $\bar x^l_t$ and their predictions $\bar \mu^l_t$ is the \emph{error} $\varepsilon^{\scriptscriptstyle {l}}_{i,t} = x^{\scriptscriptstyle {l}}_{i,t} - \mu^{\scriptscriptstyle {l}}_{i,t}$. We now describe how PCNs are trained. To do this, we explain one iteration of a training algorithm, called \emph{inference learning} (\emph{IL}), that is divided into an inference phase and a weight update phase. 

\textbf{Inference:} Only the value nodes of the network are updated, while both the weight parameters and the memory vector are fixed. Particularly, the value nodes are modified via gradient descent to minimize the global error of the network, expressed by the following energy function $E_{t}$:\vspace*{-2ex}

{\small\begin{equation}
E_{t} = 
{ \mbox{$\frac{1}{2}$}
{\textstyle\sum}_{i,l}( \varepsilon^{\scriptscriptstyle {l}}_{i,t} ) ^2}\,. 
\label{eq:pcn-f}
\end{equation}}%
%

Assume that we train a generative PCN on a training point $\bar s \in \mathbb{R}^d$. To do this, the value nodes of the sensory layer are fixed to the  training point $\bar s$, and are never updated. Thus, the error on every neuron of the sensory layer  is equal to  $\varepsilon^{\scriptscriptstyle {0}}_{i,t}  = s_{i} - \mu^{\scriptscriptstyle {0}}_{i,t}$.
%
%
The process of minimizing $E_{t}$ by modifying all $x^{\scriptscriptstyle {l}}_{i,t}$ leads to the following changes in the value nodes:\vspace*{-2ex}

{\small \begin{equation}
\!\Delta{x}^{\scriptscriptstyle {l}}_{i,t} 
= \begin{cases}
\gamma\cdot ( -\varepsilon^{\scriptscriptstyle {l}}_{i,t} + f' ( x^{\scriptscriptstyle {l}}_{i,t} ) {\textstyle\sum}_{k=1}^{n^{\scriptscriptstyle {l-1}}} \varepsilon^{\scriptscriptstyle {l-1}}_{k,t} \theta^{\scriptscriptstyle {l}}_{k,i,t} ) & \!\!\mbox{if } 0 \,{<}\ l \,\leq \ L\\
0 & \!\!\mbox{if } l=0\,,
\end{cases}
\label{eq:pcn-dotx_il}
\end{equation}}%

\noindent where $\gamma$ is the \emph{integration step}, a constant determining by how much the activity changes in each iteration. The computations in Eqs.~\eqref{eq:pcn-forward-varepsilon} and \eqref{eq:pcn-dotx_il} are biologically plausible, as they have a neural implementation that can be realized in a network with value nodes  $x^l_{i,t}$ and error nodes  $\varepsilon^l_{i,t}$ \cite{rao1999predictive}, as shown in Fig~\ref{fig:eff}. 
The inference phase works as follows: starting from a given configuration of the value nodes $\bar x_0$, inference continuously upates the value nodes according to Eq.~\eqref{eq:pcn-dotx_il} until it has converged. We call the configuration of the value nodes at convergence $\bar x_T$, where $T$ is the number of steps needed to reach convergence (in practice, it is a fixed large number).

\textbf{Weight Update:} When the value nodes of the sensory layer are fixed to an input signal $\bar s$, inference  may not be sufficient to reduce the total energy to zero. Hence, to further decrease the total error, a single \textit{weight update} is performed:
both the weight matrices and the memory vector are updated by gradient descent to minimize the same objective function~$E_{t}$, and behave according to the following equations, where $\alpha$ is the learning rate. Particularly, the derived update rule  is the following:
\begin{align}
 & \Delta \theta^{\scriptscriptstyle {l+1}}_{i,j} = -\alpha\cdot {\partial E_{T}}/{\partial \theta^{\scriptscriptstyle {l+1}}_{i,j}} 
=  \alpha\cdot \varepsilon^{\scriptscriptstyle {l}}_{i,T} f ( x^{\scriptscriptstyle {l+1}}_{j,T} ), \\
& \Delta  b_i = -\alpha\cdot {\partial E_{T}}/{\partial b_i} = -\alpha  \varepsilon^L_{i,T}.
\label{eq:pcn-update-param}
\end{align}
The phases of inference and weight update are iterated until the total energy $E_t$ reaches a minimum. This  learning algorithm  learns a dataset by using only local computations, which minimize the same energy function.  Fig.~\ref{fig:eff} gives a graphical representation of generative PCNs, while the pseudocode can be found in Alg.~\ref{algo:IL}. Detailed derivations of  Eqs.~\eqref{eq:pcn-dotx_il} and \eqref{eq:pcn-update-param} are in the supplementary material (and in~\cite{whittington2017approximation}).

\vspace*{-1ex}
\section{Predictive Coding for Associative Memories}\label{sec:fapc}\vspace*{-1ex}

So far, we have shown how PCNs can perform generative tasks. 
We now show how generative PCNs can be used as associative memories, i.e., 
how the model stores the data points that it is trained on, and how these data points can be retrieved when presenting corrupted versions to the network, 
returning the most similar stored data point.  Let $\bar s$ be a training data point, and $M$ be the PCN considered above, already trained until convergence to generate~$\bar s$. Moreover, assume that $M$ 
makes the total energy converge to \emph{zero} at iteration $T$. At this point, the energy function defined on 
the value nodes 
has a local minimum $\bar x$ in which the value nodes of the sensory layer are equal to the entries of $\bar s$. Note that $\bar x$ is actually an attractor of the dynamics of $E_t$: when given a configuration that  is not a local minimum,  inference will update the value nodes until the total energy reaches a minimum. If this configuration lies in a specific neighborhood of $\bar x$, inference will converge to $\bar x$. 
%
%
So, given a dataset, we obtain an AM of the dataset if all the training points are stored in the above way~as~attractors.

\begin{figure}[t] 
\centering
\begin{subfigure}
    \centering \hfill
    \includegraphics[width=0.47\textwidth]{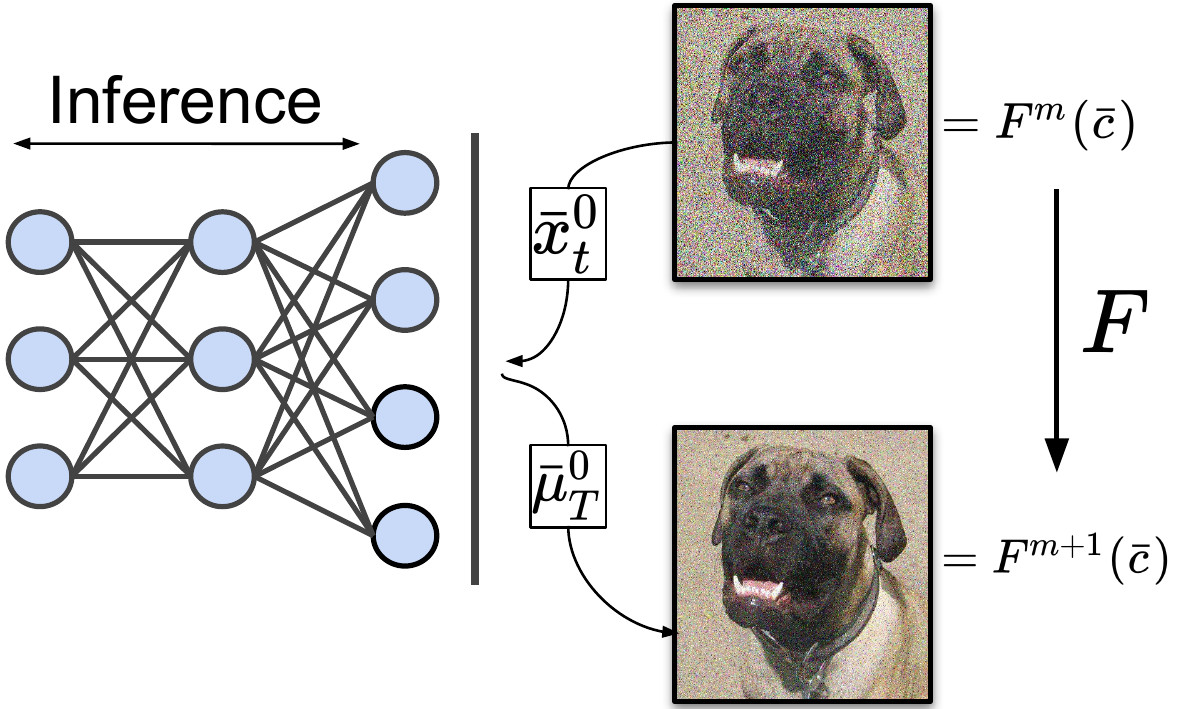}
\end{subfigure}%
\begin{subfigure}
    \centering \hfill
    \includegraphics[width=0.40\textwidth]{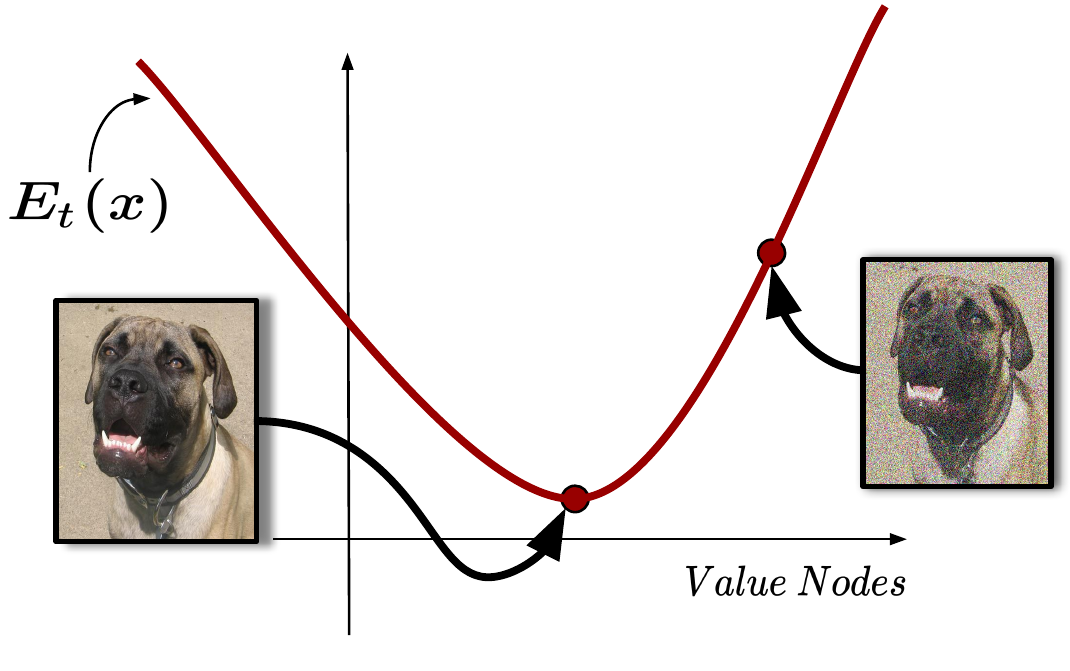}
\end{subfigure}
\vspace*{-2ex}
\caption[short]{Left: Representation of the function $F$, used to retrieve stored images. It can be decomposed into  three steps: (1) The value nodes of the sensory layer are fixed to the pixels of the corrupted image $F^m(\bar c)$. (2) Inference runs for $T$ operations (until convergence). (3) We set $F^{m+1}(\bar c)$ to  the prediction  of the sensory layer $\bar \mu_T^0$. Note that the weight parameters are never updated during the above steps. We have omitted the error nodes for simplicity. Right: Representation of an AM, where an image $\bar s$ (photo of a dog) is stored as an attractor of the dynamics. A corrupted image that lies in a specific neighborhood of $\bar s$ converges to it when minimizing the total energy via running inference.}
\label{fig:corrupted}\vspace*{-1ex}
\end{figure}

The above can be used to retrieve stored data points $\bar s$: given a corrupted version $\bar c \in \mathbb{R}^d$ of  $\bar s$, one can retrieve $\bar s$ as follows. First, we set the value nodes of the sensory layer to the corrupted points, i.e., $\bar x_t^0 = \bar c$ for the whole process.
Then, we run inference until convergence and save the prediction $\bar \mu_T^0$ of the sensory layer. If the original  data point was stored as an attractor, we expect the prediction $\bar \mu_T^0$ to be a less corrupted version of it.  %
Let $F\colon \mathbb{R}^d \,{\rightarrow}\, \mathbb{R}^d$ be the function that sends $\bar c$ to  $\bar \mu_T^0$ just described, and summarized in Fig.~\ref{fig:corrupted}. Many iterations of this function allow to retrieve the stored data point.
%
%
Hence, summarizing the above, training points are stored in the memory vector $\bar b_T$ and the weight parameters, and what the algorithm does to retrieve them is simply the inference phase of PCNs. Since visual memories are stored in hierarchical networks of brain areas, PC could be a highly plausible algorithm to better understand how memory and prediction work in~the~brain. 

\begin{figure}[t!]
\includegraphics[width=1\textwidth]{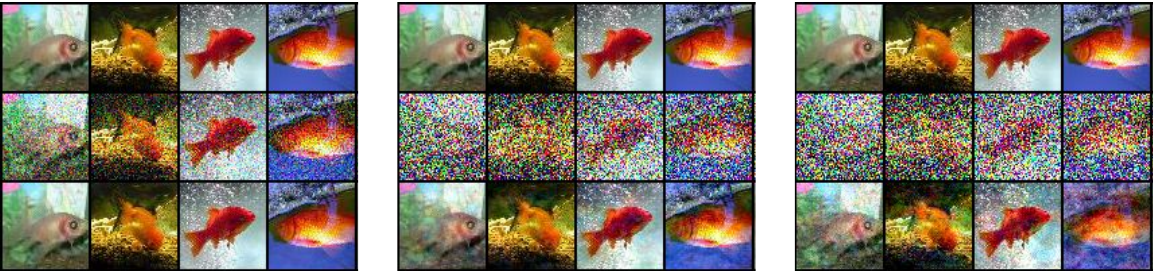}
\caption[short]{Reconstructions of $64 \times 64$ Tiny ImageNet images with noise levels $0.2$, $0.7$, and~$0.9$, respectively (original, noisy, and reconstructed image in first, second, and third row, respectively).}
\label{fig:partial_img}\vspace*{-1ex}
\end{figure}

To experimentally show that generative PCNs are AMs, we  trained a $2$-layer network with ReLU non-linearity on a subset of $100$ images of the street view house number dataset (SVHN), Tiny ImageNet, and CIFAR10. After training, we presented the model with a corrupted variant  (by adding Gaussian noise) of the training set. We  then used the PCNs to reconstruct the original images from the corrupted ones. The experimental results confirm that  the model is able to retrieve the original image, given a corrupted one. The obtained reconstructions for the Tiny Imagenet dataset (the most complex one, as each datapoint consists of $3 \times 64 \times 64$ pixels) are shown in Fig.~\ref{fig:partial_img}. We now provide a more comprehensive analysis, which studies the capacity of generative PCNs when changing the number of data points and parameters.

\begin{figure}[t!]
\begin{subfigure}
    \centering \hfill
    \includegraphics[width=1.0\textwidth]{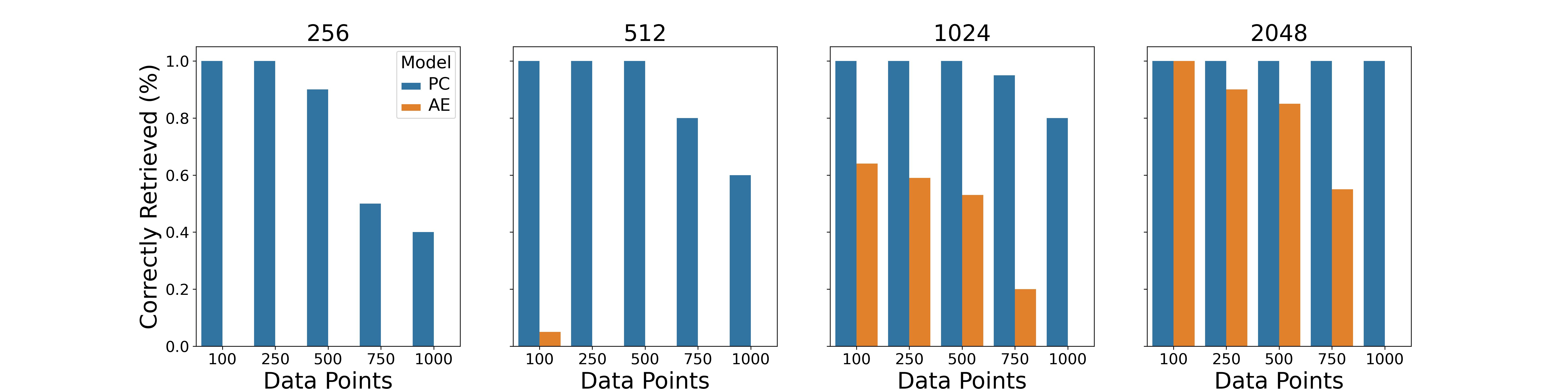}
\end{subfigure}%
\begin{subfigure}
    \centering \hfill
    \includegraphics[width=1.0\textwidth]{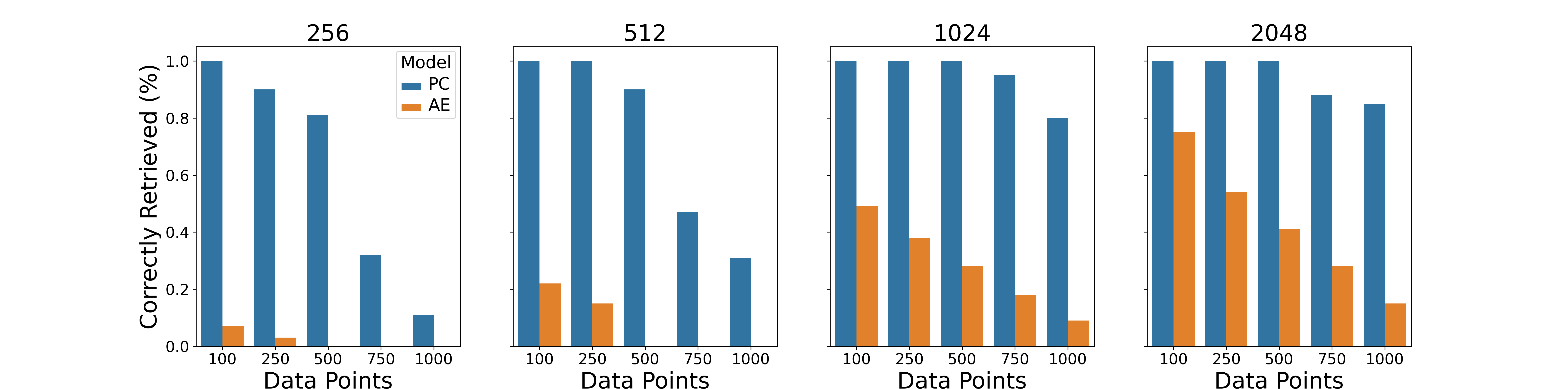}
\end{subfigure}%
\begin{subfigure}
    \centering \hfill
    \includegraphics[width=1.0\textwidth]{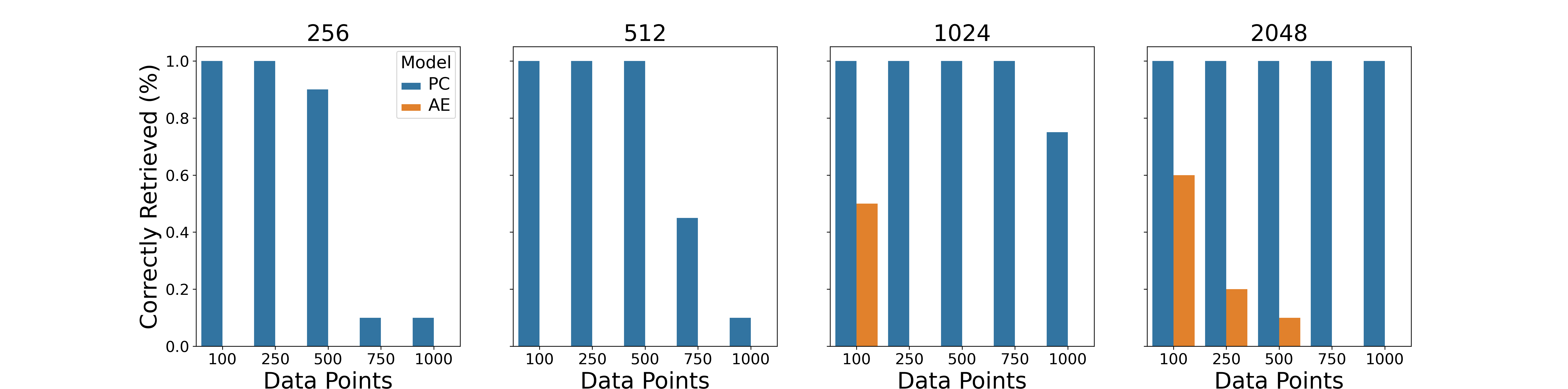}
\end{subfigure}
   \caption[short]{Percentage of correctly re\-trie\-ved images by 2-lay\-er generative PCNs (PC) and 3-lay\-er autoenco\-ders (AE) with hidden-lay\-er dimensions of 256, 512, 1024, and 2048, when presented with a corrupted image with Gaussian noise of variance $0.2$. Top row corresponds to experiments performed on CIFAR10, middle row the SVHN dataset, bottom row to Tiny Imagenet.}
\label{fig:partial_img-}\end{figure}

\textbf{Experiments:}  We trained $2$-layer PCNs with ReLU non-linearity and hidden dimension {$n \in \{256,$ $512,$ $1024,$ $2048\}$} on subsets of the aforementioned datasets of cardinality {$N=\{100,$ $250,$ $500,$ $1000,$ $1500\}$}. Every model is trained until convergence, and all the images are retrieved as explained in Section~\ref{sec:fapc}. To provide a numerical evaluation, an image is considered recovered when the error between the original image and the recovered image is less than $0.005$.

To compare our results against a standard baseline, we also trained $3$-layer  autoencoders (AEs) with the same hidden dimension on the same task, and compared the results. Note that the number of parameters of a $2$-layer PCN is smaller than the one of a $3$-layer AE with the same hidden dimension. This follows, as the additional layer (input layer, not needed in generative PCNs) almost doubles the number of parameters in some cases. Further details about the experiments and used hyperparameters are~given~in~the~supplementary material.

\textbf{Results}: The analysis shows that our model is able to store and retrieve data points even when the network is not overparametrized (see Fig.~\ref{fig:partial_img-}). AEs trained with BP did not perform well: AEs with less than $1024$ hidden neurons always failed to restore even a single data point on the Tiny Imagenet dataset, and very few on other ones.  The performance of AEs with $2048$ hidden neurons were always worse than PCNs with $256$ hidden neurons. This shows that overparametrization is essential for AEs to perform AM tasks, and that our proposed method offers a much more network-efficient alternative.

In terms of capacity, we show that $2$-layer PCNs with $256$ hidden neurons are able to correctly store datasets of $250$ images of both CIFAR10 and Tiny ImageNet, and that networks with $2048$ hidden neurons always managed to store and retrieve all the presented datasets. As typical in the AM domain, small models trained on large datasets fail to store data points, as the space of parameters is not large enough to store each data point as an independent attractor, and many attractors in the same small region lead to a chaotic dynamic. Our model is no different: PCNs with $256$ and $512$ hidden units are able to store almost $500$ Tiny ImageNet images when trained on datasets of that size, but fail to store more than $200$ when trained on larger ones ($1000$ and $1500$). 

\vspace*{-1ex}
\section{Retrieval from Partial Data Points}
\vspace*{-1ex}

So far, we have shown how the proposed generative model can be used to retrieve stored data points when presented with corrupted variants. We now tackle the different task of retrieving data points when presented with partial ones. 
Let $\bar s$ be the stored data point, and assume that a fraction of pixels of $\bar s$ are accessible, and the goal is to retrieve the remaining ones. Let  $\bar s'$ be the vector of the same dimension of $\bar s$, where a fraction of pixels are equal to the ones of the stored data point, and assume that the position of the pixels that are equal to the ones of $\bar s$ is known. We now show how to retrieve the complete data point by using the same network~$M$, trained as already shown in Section~\ref{sec:fapc}.

\begin{figure}[t]
\centering
    \includegraphics[width=0.65\textwidth]{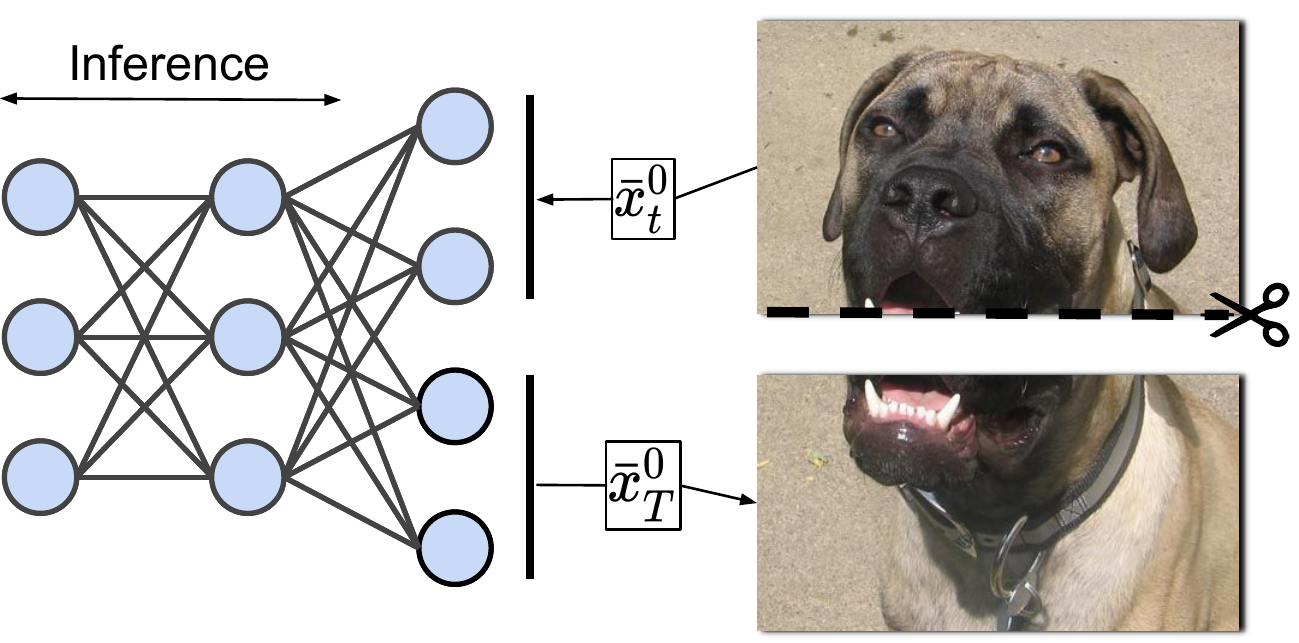}
\caption[short]{Algorithm used to retrieve a stored data point from a fraction of it: 
(1) The value nodes of the sensory layer that correspond to the available entries of $\bar s$ are fixed to the respective values. (2) Inference runs for $T$ operations (until convergence). (3) We set our prediction to be equal to the value nodes of the sensory layer $\bar x_T^0$. The weight parameters are never updated during the above steps. The error nodes are omitted for simplicity. }
\label{fig:partial}
\vspace*{-2ex}
\end{figure}

Let $M$ be a PCN trained to generate $\bar s$. Then, given the partial version of the original data point $\bar s'$, it is possible to retrieve the full data point using $M$ as follows: first, we only fix the value nodes of the sensory layer $\bar x_t^0$ to the entries of the partial data point $\bar s'$ that we know are equal to the ones of the stored data point, leaving the rest free to be updated. Then, we run inference on the whole network until convergence. At this point, we expect the value nodes $\bar x_T^0$ to have converged to the entries of the original data point, stored as an attractor. A graphical representation of the above mechanism is described in Fig.~\ref{fig:partial}. To show the capabilities of this network, we have performed multiple experiments on the Tiny ImageNet and ImageNet datasets, and compared against existing models in the literature. We now start by providing visual evidence on the effectiveness of this method. Note that the geometry of the mask does not influence the final performance, as our model simply memorizes single pixels.

\textbf{Experiments:} We trained two networks with hidden dimensions of $1024$ and $2048$, to generate $50$ images of the first class of Tiny ImageNet (corresponding to goldfishes), and a network of $8192$ hidden neurons to reconstruct $25$ pictures taken form ImageNet. Then, we used inference as explained to retrieve the original images. We considered an image to be correctly reconstructed when the error between the original and the retrieved image was smaller than $0.001$. Furthermore, we plotted the partial images together with their reconstructions, for a visual check. Note that we have used the thresholds that provided the fairest comparison: the denoising experiments fail to have a perfect retrieval, despite the fact that most of the images look visually good. Hence, we have determined the threshold to be equal to 0.005. Then, with the same threshold for the retrieval of partial images, our method always successfully retrieved all the images, and so we have opted for a smaller threshold, which was more informative.

\begin{algorithm}[t]
    \caption{Retrieving $\bar s$ given a non corrupted fraction $\bar s'$}\label{algo:IL_partial}
    \begin{algorithmic}[1]
    \REQUIRE if $s'_i$ a correct entry of the original memory, then $x^0_{i,t}$ is fixed to  $s'_i$.
    \FOR{$t=1$ to $T$}
        \FOR{each neuron $i$ in each level $l$}
            \STATE Update $x^{\scriptscriptstyle {l}}_{i,t}$ to minimize $E_{t}$.
        \ENDFOR
    \ENDFOR
    \STATE \textbf{return} $\bar{x}^0_T$
    \end{algorithmic}
\end{algorithm}

\begin{figure}[t]
\includegraphics[width=1\textwidth]{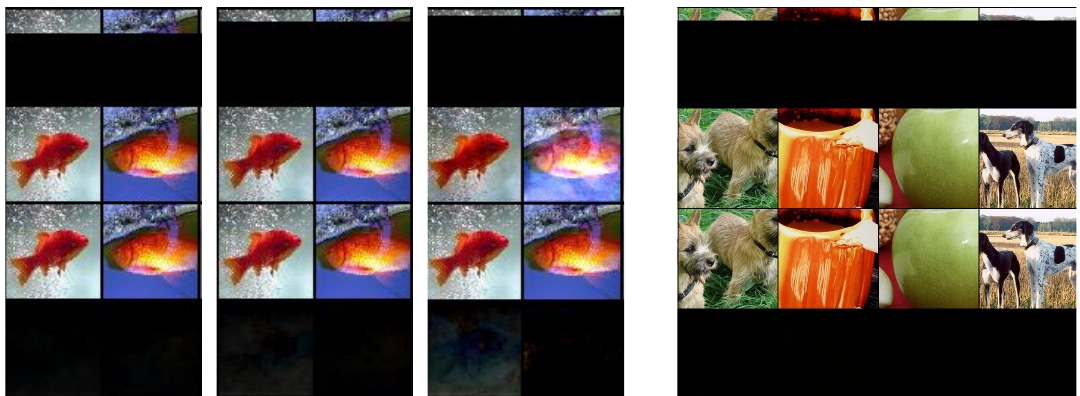}
\hfill 
\vspace*{-2ex}
\caption[short]{ Partial images and their reconstructions (from top to bottom: partial image, reconstructed image, original image, and reconstruction error (difference between original and reconstruction)). Left: $64 \times 64$ Tiny ImageNet images presented with $1 /4$, $1 /8$, and $1/ {16}$ of the original image, respectively, using a network of $1024$ hidden neurons.  Right: $224 \times 224$ ImageNet images presented with $1/ 8$ of the original image, using a network with 8196 hidden neurons.  }
\label{fig:long-img}\vspace*{-2ex}
\end{figure}

\textbf{Results:} On Tiny ImageNet, a generative PCN with $1024$ hidden neurons managed to reconstruct all the images when presented with $ 1/ {8}$ of the original image, and more than half for the smallest fraction considered, $1/16$. The network with $2048$ neurons also failed to reconstruct all the images when presented with $ 1/ {16}$ of the original image. However, even when presented with a portion as small as $1 /{16}$ of the original image, the reconstruction was clear, although not perfect and hence now aove our threshold. This result is shown in Fig.~\ref{fig:long-img} (left).

Surprisingly, generative PCNs trained on ImageNet correctly stored all the presented training images, and correctly reconstructed them with no visible error. This shows that PCNs can be used to store high-dimensional and high-quality images in practical setups, which can be retrieved using only a low-dimensional key, formed by a fraction of the original image. 
Particularly, Fig.~\ref{fig:long-img} (right) shows the perfect reconstruction obtained on ImageNet when the network is presented with only $1/8$ of the original pixels. Further experiments on ImageNet are shown in the supplementary material, where we present multiple high-quality reconstructions.

\vspace*{-1ex}
\section{More Training Data Points and/or Deeper PCNs}\vspace*{-1ex}

\begin{figure}[t]
\centering

    \includegraphics[width=1\textwidth]{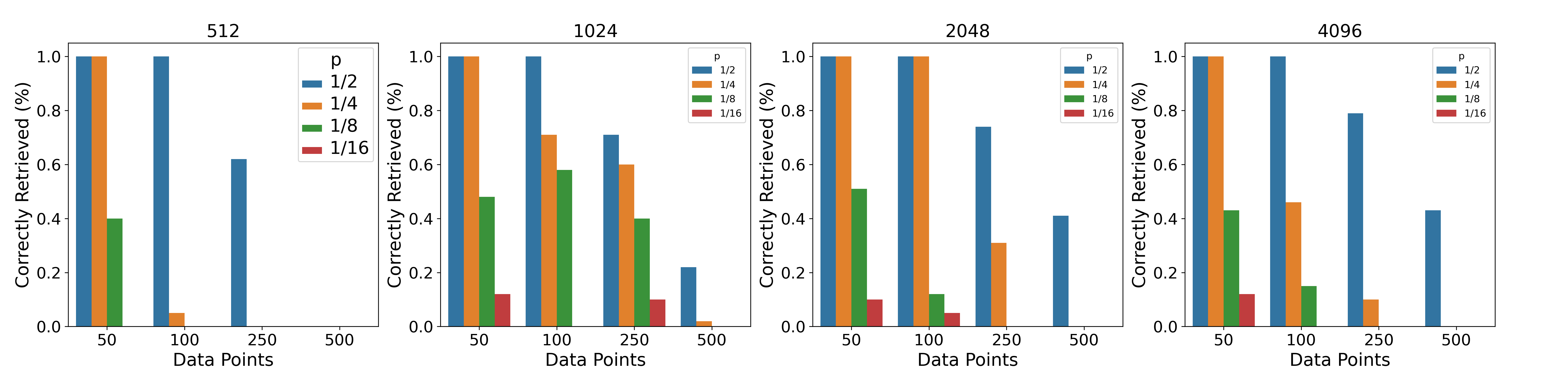} 
\caption[short]{Percentage of perfectly retrieved images  given different fractions of the original images for PCNs with $512$, $1024$, $2048$, and $4096$ hidden neurons trained on datasets of different~sizes. Note that here we count as retrieved only images with non-visible errors. Most of the images that are not considered retrieved in this plot, were correctly retrieved, but only slightly noisy. }
\label{fig:edge_plots}\vspace*{-2ex}
\end{figure}

In the above experiments, a shallow architecture is able to store $100$ natural images from  Tiny ImageNet, and to reconstruct them with no visible error when provided with $1/8$ of the original images. We now study how this changes when either the cardinality of the dataset is increased, or the number of provided pixels during reconstruction is updated. To study this, we have trained two models with $n=\{1024,2048\}$ hidden units on subsets of Tiny ImageNet of cardinality $N=\{50,100,250,500\}$, and reconstructed images with only $p=\{ 1/2,1/4,1/8,1/16\}$ of the original pixels. According to our visualization experiments, we have noted that an image has no visible reconstruction noise when the error between the original image and the reconstruction is less than $0.001$. Hence, we consider an image to be correctly retrieved if the reconstruction error is below this threshold.

\textbf{Results:}  Every network with $50$ stored memories was able to perfectly reconstruct the original image when provided with $1/4$ of the original image, and about half when provided with only $1/8$. However, this changes when increasing the training set, as no network trained on more than $50$ images was able to correctly reconstruct an image when provided with $1/16$ of the original images, besides a few cases. In terms of capacity, the more images we train on, the harder the reconstruction becomes: no training image was correctly retrieved when training on $N=500$ images when provided with a fraction smaller than $1/2$ of the original image, regardless of the width of the network. A summary of these results is given in Fig.~\ref{fig:edge_plots}.

These experiments show the limits of shallow generative PCNs on both reconstruction capabilities and capacity. As expected and common in standard AM experiments, increasing the number of training points and reducing the number of available information for reconstruction hurts the performance. We now show how increasing the number of layers solves this capacity problem.
\vspace*{-0.5ex}
\subsection{Deep Generative PCNs}\vspace*{-0.5ex}

We now test how increasing the depth of generative PCNs increases their capacity. We then compare the results against overparametrized AEs. Particularly, we trained generative PCNs with~$n = \{1024,$ $2048\}$ hidden neurons, depth $L=\{3,5,7,10\}$ on $N=500$ images of Tiny ImageNet. Furthermore, we reconstructed the stored images using a fraction of $p=\{1/2,1/4\}$ of the original pixels. To further compare against state-of-the-art AEs, we  trained equivalent AEs, which is the best-performing one according to \cite{radhakrishnan19}. Also, all the hyperparameters used for training are the ones reported by the authors. To make the comparison completely fair, we also assume that the correct pixels of the incomplete images are known when testing the AE. Particularly, we  fixed these pixels  at every iteration of the reconstruction process of the AEs. As above, we consider an image to be correctly retrieved if the reconstruction error is less than $0.001$.

\begin{figure}[t]
\centering
\includegraphics[width=1\textwidth]{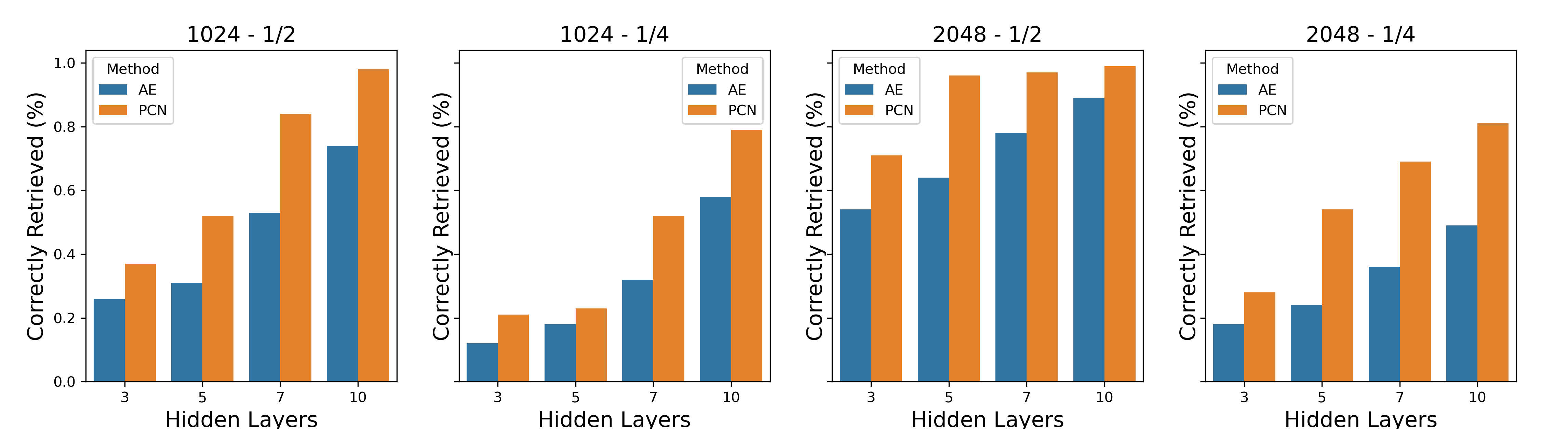}

\caption[short]{
Percentage of correctly retrieved images as a function of the number of hidden layers, for both AEs and generative PCNs, on a dataset of $500$ images. We used networks with $n=\{1024,2048\}$ hidden neurons, and the images are taken from  Tiny ImageNet.}
\label{fig:deep_plots}\vspace*{-2ex}
\end{figure}

\textbf{Results:} The results confirm the hypothesis: the deeper the network, the higher the capacity. Particularly, a $10$-layer network was able to reconstruct more than $98\%$ (against the $72\%$ of the AE) of the images when providing half of the original pixels, and $74\%$ (against the $48\%$ of the AE) of the images when providing $1/4$ of the pixels. This shows that our model clearly outperforms state-of-the-art AEs in image reconstruction, even in the overparametrized regime. Fig.~\ref{fig:deep_plots} summarizes these results.

\section{Comparison with Modern Hopfield Networks}

Hopfield networks are the most famous AM architectures. However, they have a limited capacity and only work with binary datasets. To solve these limitations, modern Hopfield networks (MHNs) were introduced \cite{ramsauer21,Krotov21}. There are significant differences between MHNs and our generative PCNs, which strongly influence the possible application domains. MHNs are one-shot learners, and can store exponentially many memories. Hence, training an MHN is faster than training a generative PCN. Furthermore, they are able to exactly retrieve data points, while our model always presents a tiny amount of error, even if not visible by the human eye. However, the retrieval process of our model is significantly better, as it always converges to a plausible solution, even when provided with a tiny amount of information, or a large amount of corruption. For example, our model never converges to wrong data points: when tested on complex tasks, it simply outputs fuzzier memories, instead of perfect but wrong reconstructions, as the MHNs. We now use MHNs to perform image retrieval and reconstruction experiments, and compare the results against generative PCNs. 

\textbf{Experiments:} We have trained an MHN to memorize 50/100/250/500 images of the CIFAR10 dataset, and then performed three different experiments. In the first two, we tried to reconstruct the original memories when providing $1/ 2$ and $1 /4$ of the original image. In the third experiment, we have corrupted the original data points by adding Gaussian noise  $\eta= 0.2$.

\begin{figure}[t]
\centering

    \includegraphics[width=1\textwidth]{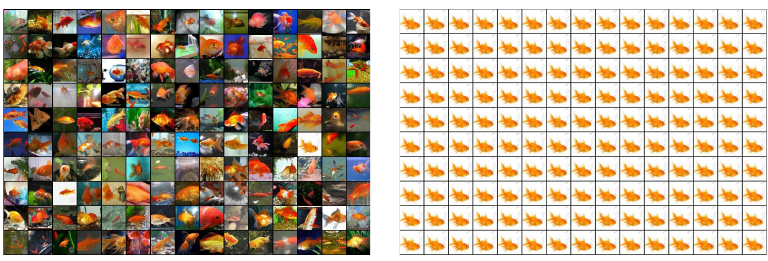}

\caption[short]{Left: Reconstruction of $150$ images of the Tiny ImageNet dataset when provided with $1/2$ of the original image. To generate the figure, we have used a generative PCN with $2$ hidden layers and hidden dimension $n=1024$. Right: Same task performed using an MHN with $\beta = 2$. Overall, our method has retrieved all the presented images, while the MHN has retrieved only one, which seems to correspond to the only strong attractor of the dynamics.}
\label{fig:MHN}\vspace*{-2ex}
\end{figure}

\textbf{Results:} The results are summarized in Table~\ref{tab:MHN}. Despite the high capacity, MHNs do not perform well in image retrieval. Particularly, they were able to retrieve at most $9$ images when presented with a corrupted data point, and at most $5$ when presented with an incomplete one. The vast majority of the reconstructions correspond to wrong images of the original training set. However, every correctly retrieved image is an exact copy of the original memory. This shows that only a few data points are stored as strong attractors. In Fig.~\ref{fig:MHN}, we provide a visual comparison between our method and MHNs, showing that, when images are more complex, MHNs are only able to retrieve one image.

To perform the experiments on MHNs, we have used the official implementation, provided by the authors and available online. We now provide further details about the experiments, sufficient to replicate the results.

One of the qualities of MHNs, is that they only depend on one hyperparameter, $\beta$, as the hidden dimension is equivalent to the number of data points. This parameter was extensively tested, as we have used $\beta \in \{1,2,3,5,10,100,1000\}$, and always reported the best result. Furthermore, to make our analysis more fair, we have also stored the images multiple times. In fact, in the reported numbers, the hidden dimension of the network is also provided: a network with $3N$ hidden neurons trained on $100$ images indicates that we have stored each image $3$ times.

\begin{table}
\footnotesize
\begin{tabular}[t]{cccc}
    \toprule
    \cmidrule(r){1-4}
     Hidden Dimension & $N$ & $ 3N$ & $5N$          \\
    \midrule
    50      & 5 & 5 & 4  \\
    100      & 4 & 4 & 4   \\
    250      & 1 & 1 & 1  \\
    500      & 1 & 1 & 1   \\
    \bottomrule
    \end{tabular}
\hfill
\begin{tabular}[t]{cccc}
    \toprule
    \cmidrule(r){1-4}
     Hidden Dimension & $N$ & $ 3N$  & $5N$                \\
    \midrule
    50      & 4 & 3 & 3  \\
    100      & 3 & 3 & 3   \\
    250      & 1 & 1 & 1  \\
    500      & 1 & 1 & 1   \\
    \bottomrule
    \end{tabular}
\hfill
\begin{tabular}[t]{cccc}
    \toprule
    \cmidrule(r){1-4}
     Hidden Dimension & $N$ & $ 3N$  & $5N$                \\
    \midrule
    50      & 5 & 4 & 4   \\
    100      & 2 & 3 & 2   \\
    250      & 8 & 9 & 8  \\
    500      & 3 & 3  & 3  \\
    \bottomrule
    \end{tabular}
    \vspace{0.3cm}
        \caption{Number of correctly retrieved images by an MHN trained on $N=\{50,100,250,500\}$ images of the CIFAR10 dataset on different tasks. Left: The model was provided with $1/2$ of the original image. Centre: The model was provided with $1/4$ of the original image. Right: The model was provided with a corrupted version of the original image, with injected Gaussian noise $\eta = 0.2$.}
        \label{tab:MHN}
        \end{table}

\vspace*{-0.5ex}
\subsection{Comparison with Deep Associative Neural Networks}\vspace*{-0.5ex}

As shown, the original formulation of MHNs does not allow to perform image reconstruction experiments. To solve this problem, a new model has been presented, which consists of a MHN augmented with a convolutional multilayer structure that performs unsupervised feature detection \cite{LIU19} using a deep belief network. The resulting architecture, called \emph{deep associative neural network} (DANN), is not a pure associative memory model, as it is able to perform both AM and classification tasks. We now compare the image reconstruction experiments performed by the authors, with the ones performed by our generative PCN. We replicate the experiment proposed, which consists in presenting the network a partial image of the CIFAR10 dataset, where a squared patch covers the centre of each image. 

Note that the comparison we propose is purely qualitative, as DANNs auguments MHNs with an unsupervised feature detection, using convolutions to further improve their results, and are  both memorizing the images and learning the main feature patterns. This allows them to have a large capacity, which allows DANNs to be trained on large datasets. Hence, comparing it against a pure AM model such as ours may not be indicative. However, we believe that presenting comparisons against models that perform AM tasks such as DANNs is still interesting in understanding how our model compares against existing ones in the literature.

\begin{figure}[t]
\centering

    \includegraphics[width=1\textwidth]{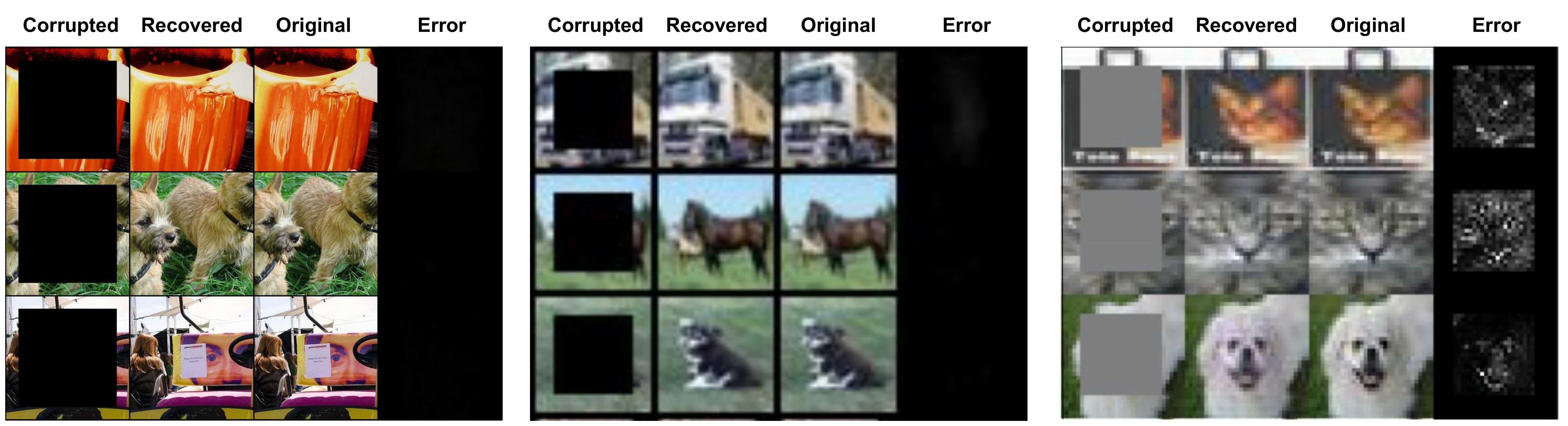}

\caption[short]{Left: Our reconstruction on $50$  ImageNet $224{\times} 224$ images. Centre:  Our reconstruction using a generative PCN with 1024 hidden neurons on a fraction of the CIFAR10 dataset. Right: Reconstruction of DANN on the whole CIFAR10 dataset (taken from \cite{LIU19}).}
\label{fig:DANN}\vspace*{-2ex}
\end{figure}

\textbf{Results:} We first ran the same experiment on CIFAR10 as in \cite{LIU19}. Particularly, we  used $500$ images. The comparison is given in Fig.~\ref{fig:DANN}. Unlike the image in \cite{LIU19}, our reconstruction shows no visible error. The results show that the reconstruction of our AM model are much clearer than the ones in \cite{LIU19}, and so consistently improving over this particular task. Furthermore, we replicated the experiment using $50$ images of ImageNet, and showed that our model is again able to provide perfect reconstructions of the original images. However, DANNs show a larger capacity, as the provided images are obtained after training on the whole CIFAR10 dataset.

\section{{Hetero-Associative Memory Experiments}}

Generally speaking, associative memories come in two high-level forms: auto-associative and hetero-associative memories. While both are able to retrieve memories given a set of inputs, auto-associative memories are primarily focused on recalling a specific pattern when provided a partial or noisy variant of X. So far, we have only performed this kind of task. By contrast, hetero-associative memories are able to recall not only patterns of different sizes from their inputs, but are also able to memorize multi-modal information, such as associate natural language with sounds or images. In this section, we study the performance of this model with respect to multi-modal dataset, by training it on images and their textual descriptions, and then using the description to retrieve the related image.

\textbf{Experiment:} The dataset used for this task is \texttt{flickr\_30k} \cite{flickr30k}, which consists of colored captioned images of varying size, that we resize and reduce to $32\times 32$ colored images. Each image has 5 associated captions, from which we only use the first caption. The captions are tokenized by first building a vocabulary (using the package \texttt{spacy}) and then discretized using said vocabulary. Afterwards, they are transformed into floating point numbers and normalized to $[0,1]$, to remain in the same range as the images. Finally, the captions are padded with $0$s to the size of the longest caption, to maintain the same data size. For example, the captions associated with the first three images are:
\begin{enumerate}
    \item Two young guys with shaggy hair look at their hands while hanging out in the yard;
    \item Several men in hard hats are operating a giant pulley system;
    \item A child in a pink dress is climbing up a set of stairs in an entry way.
\end{enumerate}

Note that the task of retrieving images from captions is complex, as every pixel of an image image of size  $3072$ has to be retrieved using a key of dimension $25$ (hence, using $<1\%$ of the original information). 

To show how depth influences the retrieval quality, we have trained different PCNs with depth $L\in\{1,3,5,7,9,11\}$ and width $n\in \{512, 1024, 2048\}$ to memorize $25$, $50$, and $100$ captioned images. Then, the networks are given the images without to recover the captions and vice-versa. The number of recovered captions and recovered images are then recorded. An image is considered to be recovered if its error is less than $0.001$, while a caption is considered to be recovered if it is \emph{exactly} recovered: unlike pixels, words are embedded as discrete information, and hence a word is either exactly retrieved, or mistaken by a wrong word of the dictionary. A caption is considered to be correctly recovered if every each word is correctly recovered.

\begin{figure}[!tbp]
    \centering
    \begin{minipage}[b]{0.49\textwidth}
      \includegraphics[width=\textwidth]{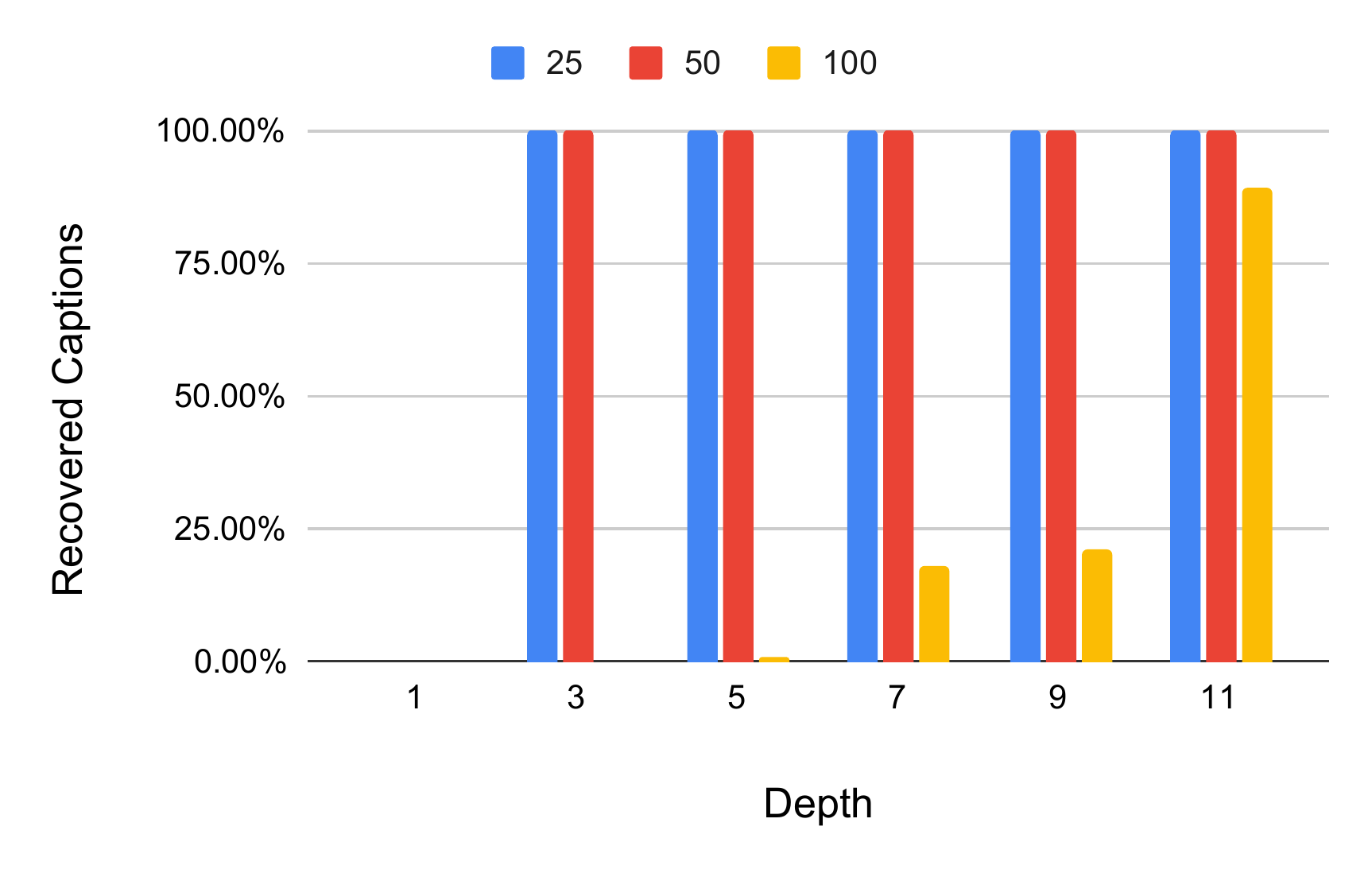}
    \end{minipage}
    \hfill
    \begin{minipage}[b]{0.49\textwidth}
      \includegraphics[width=\textwidth]{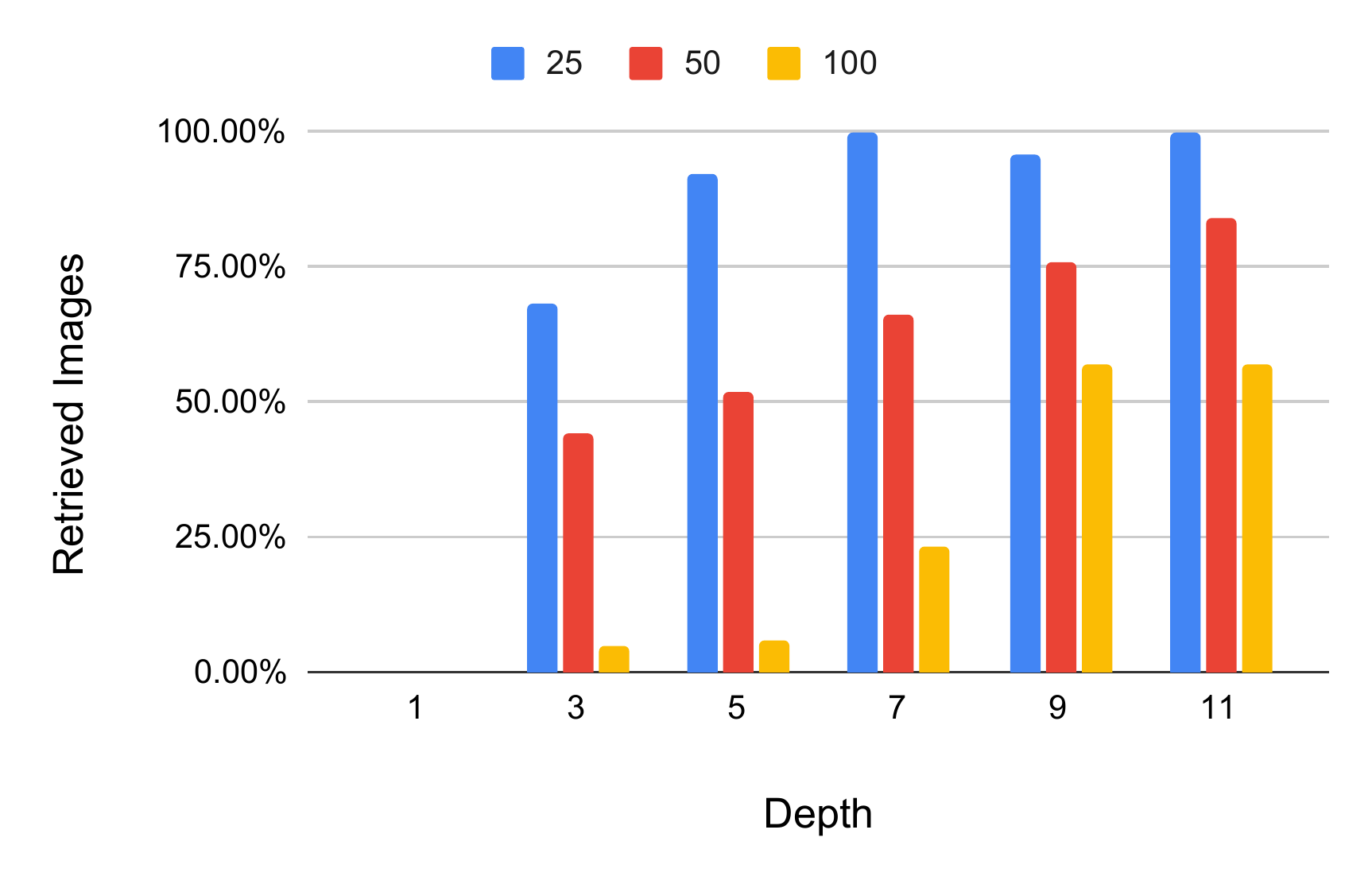}
    \end{minipage}
    \begin{minipage}[b]{0.49\textwidth}
      \includegraphics[width=\textwidth]{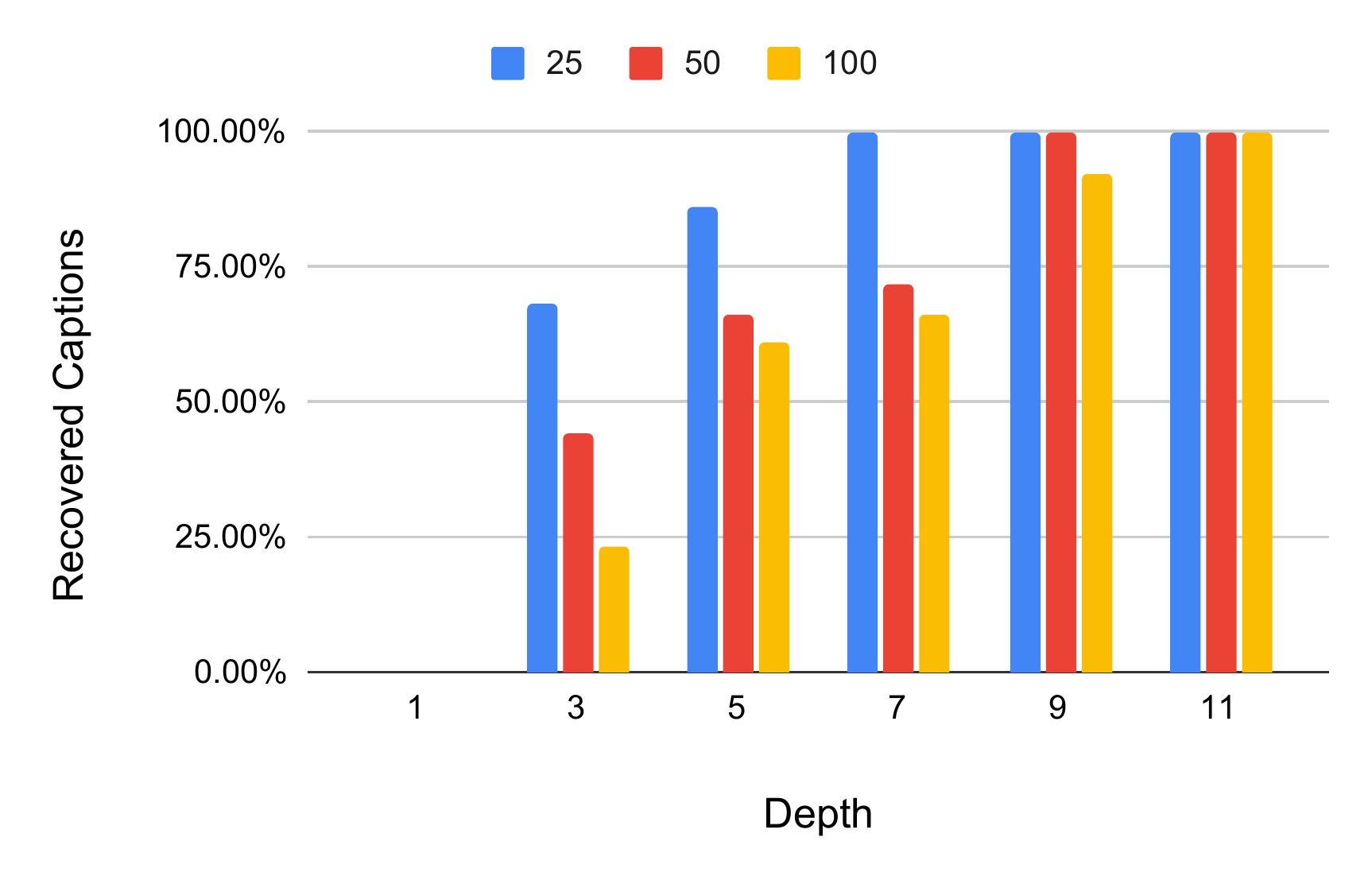}
    \end{minipage}
    \hfill
    \begin{minipage}[b]{0.49\textwidth}
      \includegraphics[width=\textwidth]{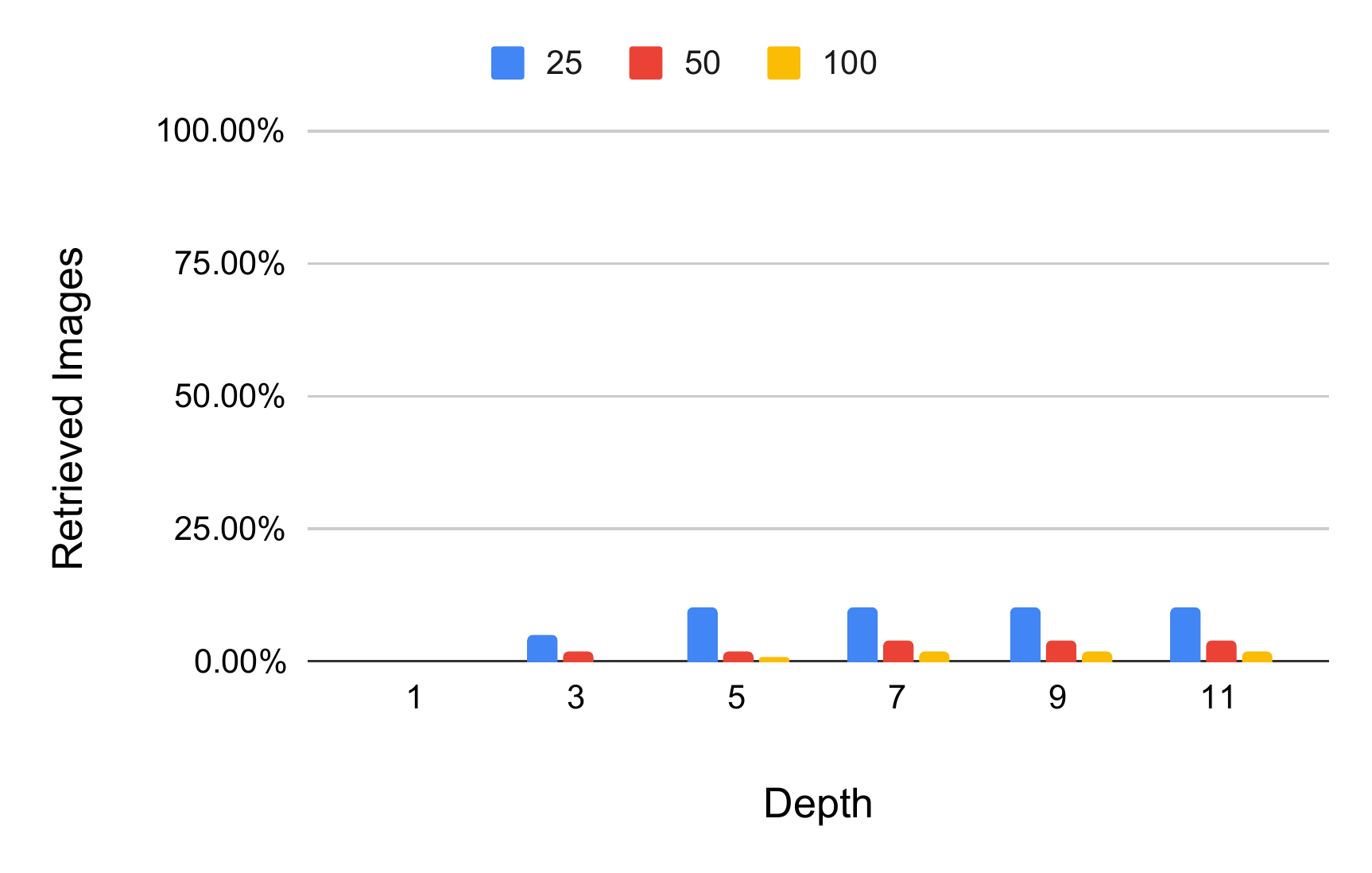}
    \end{minipage}
    \caption[Number of recovered captions and images for generative PCNs of different depths.]{Depths comparison for generative PCNs with respect to hetero-associative memories. The top row shows the results for generative PCNs, and the bottom row the ones for AEs. The number of captions (resp., images) recovered is shown on the left (resp., right) side. Both models are trained for the depths $L\in\{1,3,5,7,9,11\}$ and the widths $n\in\{512,1024,2048\}$, with the best number across $n$ being shown.}
    \label{fig:ha_comparison_depth}
\end{figure}

\textbf{Results:} The number of images and captions recovered are given in Fig.~\ref{fig:ha_comparison_depth}. We found that increasing the number of neurons in each layer did not affect the results much, then only the best result from each depth is shown. The results indicate that increasing the number of layers in the network increases the capacity of the network, as expected. Most of the networks successfully recovered most of $25$ images from the respective captions, which means that from $\simeq 1\%$ of the original data the networks remembered the remaining $99\%$ and more than half of the images when trained on $50$ datapoints. When trained on $100$ datapoints, only deep networks ($9$ and $11$ hidden layers) were able to perfectly retrieve more than half of the datapoints. The shallow networks of depth 1 performed poorly on all the tasks, not recovering any image or caption. Compared to AEs, PCNs perform slightly worse when retrieving captions, but significantly better when retrieving pictures, as AEs were never able to retrieve more than two, regardless of the parametrization or the number of data points used.

\begin{figure}[t]
    \centering
    \begin{minipage}[b]{0.35\textwidth}
      \includegraphics[width=\textwidth]{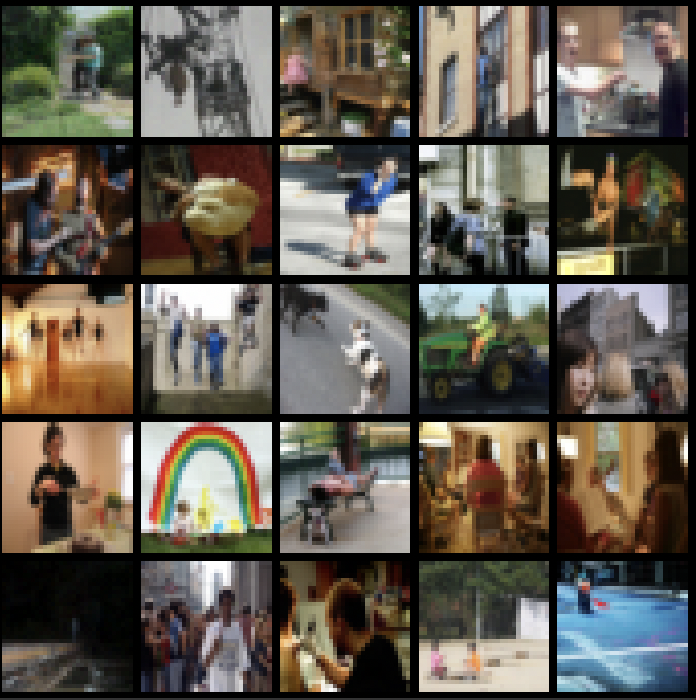}
    \end{minipage}
    \hspace{1em}
    \begin{minipage}[b]{0.35\textwidth}
      \includegraphics[width=\textwidth]{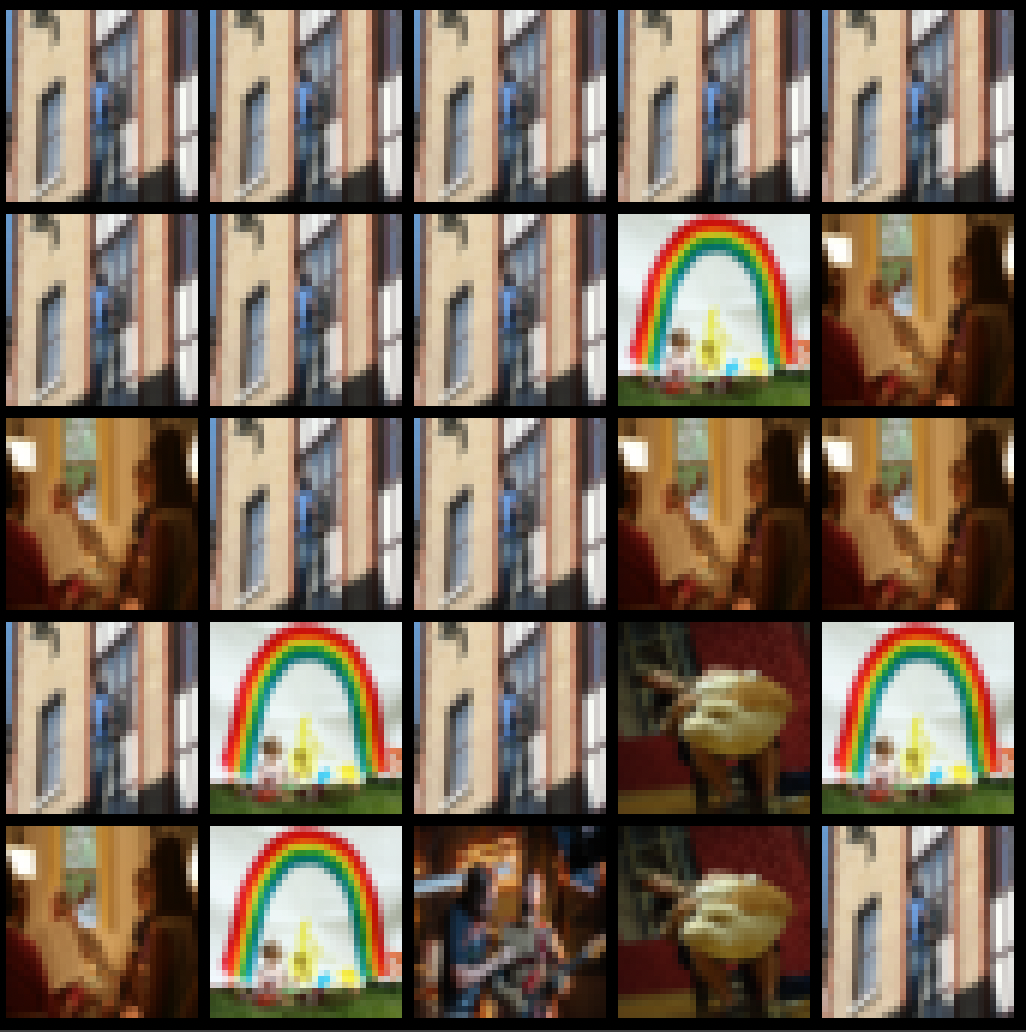}
    \end{minipage}
    \caption[]{Recovery of 25 colored images from captions.  Left: images recovered by a generative PCN. Right: images recovered by an AE with the same number of parameters. 
    For generative PCNs, all images are visually identical to the original ones. For AEs, only 2 images are correctly recovered from captions, although  5 images are stored.
    }
    \label{fig:bp_recover_25}
\end{figure}

The results show that generative PCNs are able to both handle multimodal data, and retrieve stored memories using less than $1\%$ of the original information. Furthermore, an analysis of the retrieved images of the AEs, show that they are always able to correctly retrieve images of the dataset, even when it does not correspond to the one we were looking for. This shows that AEs store images as attractors, but some attractors are stronger than others, and hence most of the retrieval processes converge to the unrelated memories. This result is visually shown in  Fig.~\ref{fig:bp_recover_25}, which represents the recovery of 25 images by both PCNs and AEs.

\section{{Biological Relevance}}

The model described in this paper is closely related with theories of functioning of the memory system in the brain. It is typically assumed that memories are stored in a network of cortical areas organized hierarchically, where sensory information is processed and encoded by a several neo-cortical regions, and then it is integrated across modalities in the hippocampus \cite{Felleman91,Mcclelland95}. It has been recently proposed that this hippocampo-neocortical system can be viewed as a predictive coding network \cite{Barron20} similar to that simulated in this paper. In particular, Barron et al. \cite{Barron20} proposed that the hippocampus could be viewed as a top layer in the hierarchical predictive coding network (which would correspond to the memory layer $L$ in our model), while neo-cortical areas would correspond to lower layers in the network. A~contribution of our paper is to implement this idea and demonstrate in simulations that predictive coding networks, which were only theoretically proposed to act as a memory system \cite{Barron20}, can actually work and effectively store and retrieve memories.



The mechanisms of learning and retrieval proposed in the predictive coding account of the brain memory system in \cite{Barron20} closely resemble those occurring in our model.
Barron et al. proposed that during learning, the generative model of the hippocampus is updated until the hippocampal predictions correct the prediction errors by suppressing neo-cortical activity \cite{Barron20}. The training phase of our generative model closely resembles this framework, as it also accumulates prediction errors from the sensory neurons to the hidden layers and memory vector, while a continuous update of the value nodes corrects the errors in the sensory neurons.  Then, the parameters of our generative model are updated until its predictions reach low/zero error in the sensory neurons. Learning in predictive coding networks relies on local Hebbian rules, i.e., the weights between co-active neurons are increased, and such form of synaptic plasticity has been first demonstrated experimentally in the hippocampus \cite{bliss1973long}.
It has been further suggested that during the reconstructions of past memories, the hippocampus sends descending input to the neocortical neurons to reinstate activity patterns that recapitulate previous sensory experience \cite{Barron20}. In the reconstruction phase of our generative model, the memory vector provides descending inputs to the sensory neurons, to generate patterns that recall previous sensory experience, e.g., stored~data points.


Our model captures certain high-level features of the brain memory system, but many details of the model will have to be refined to describe biological memory networks. Barron et al.\ \cite{Barron20} reviewed in detail how the architecture of predictive coding model could be related to the known anatomy of neo-cortical memory areas, and how different populations of neurons and connection in the model could be mapped on known groups of neurons and projections in the neo-cortex. Nevertheless, it would be an interesting direction of future work to extend the top layer of our memory network to include key features of the hippocampus, i.e., recurrent connections, known to exist in the subfield CA3, as well as an additional population of neurons in dentate gyrus and CA1, which have been 
proposed to play specific roles in learning new memories \cite{oreilly94}. It will be also interesting to investigate if our model can reproduce experimental data on how the hippocampus drives the reinstantiation of stored patterns in the neo-cortex during retrieval \cite{cowansage14,tanaka14}.

\vspace*{-1ex}
\section{Related Work}\vspace*{-1ex}

Predictive coding \cite{rao1999predictive} and classic AM models \cite{Hopfield82,Hopfield84} have always been unrelated topics in the literature. However, they share multiple similarities, the main of which is that both are energy-based models that update their parameters using local information.

\textbf{Associative Memories (AMs): } In computer science, the concept of AMs dates back to $1961$, with the introduction of the \emph{learn matrix} \cite{Steinbuch61}, a hardware implementation of hetero-associative memories using ferromagnetic properties. However, the milestones of the field are Hopfield networks, presented in the early eighties in their discrete \cite{Hopfield82} and continuous \cite{Hopfield84} version. Recently, AMs have increasingly gained popularity. Particularly, a $2$-layer version of Hopfield  networks with polynomial activations has been shown to have a surprisingly high capacity \cite{Krotov16}. This capacity can be increased even more when having exponential activations \cite{Demircigil17,Krotov21}. To improve the retrieval process of stored memories, different energy-based models have been developed, such as \cite{bartunov2019}

The recent focus of the machine learning community on AM models, however, does not lie in applying them directly to solve practical hetero-associative memory tasks. On the contrary, it has been shown that modern architectures are implicitly  AMs. Examples of these are standard deep neural networks \cite{Poggio21AM,radhakrishnan19} and transformers \cite{ramsauer21}. Hence, there is a growing belief that understanding AM models will help the understanding and improvements of deep learning \cite{widrich20,Poggio21AM,domingos20,Feldman20}. In fact, different classification models have been shown to perform well when augmented with AM capabilities. DANNs \cite{LIU19}, e.g., use deep belief networks to filter information, while \cite{YANG20} improves over standard convolutional models by adding an external memory that memorizes associations among images. 

\textbf{Predictive Coding (PC): } PC is an influential theory of cortical function in both theoretical and computational neuroscience. It is unique in the sense that it provides a computational framework, able to describe information processing in multiple brain areas \cite{friston2005theory}.
It has different theoretical interpretations, such as free-energy minimization~\cite{bogacz2017tutorial,friston2003learning,friston2005theory,whittington2019theories} and variational inference of  probabilistic models~\cite{whittington2017approximation}.
Furthermore,  its single mechanism is able to account for different perceptual phenomena observed in the brain. Among them, we can list end-stopping~\cite{rao1999predictive}, repetition-suppression~\cite{auksztulewicz2016repetition}, illusory motions~\cite{lotter2016deep,watanabe2018illusory}, bistable perception~\cite{hohwy2008predictive,weilnhammer2017predictive}, and attentional modulation of neural activity~\cite{feldman2010attention,kanai2015cerebral}.

Due to this solid biological grounding, PC is also attracting interest in the machine learning community recently, especially focusing on finding the links between PC and BP \cite{whittington2017approximation,millidge2020predictive}. From the more practical point of view, PC can be used to train high-performing neural architectures \cite{whittington2017approximation,millidge2020predictive}, as it has been shown that it can exactly replicate the weight update of BP \cite{Song2020,Salvatori2021,salvatori2021any}.

\vspace*{-1ex}
\section{Conclusion}\vspace*{-1ex}
In this paper, we have shown that predictive coding (which was originally designed to simulate learning in the visual cortex) naturally implements associative memories that 
have a high retrieval accuracy and robustness, and that can be expressed using small and simple fully connected neural architectures. This has been extensively shown in a large number of experiments, for different architectures and datasets. Particularly, we have performed denoising and image reconstruction tasks, comparing against popular AM frameworks. Furthermore, we have shown that this model is able to reconstruct with no visible error  natural high-resolution images of the ImageNet dataset, when provided with a tiny fraction of them. From the neuroscience perspective, we have shown that our model closely resembles the ``learn and recall" phases of the hippocampus.

Furthermore, this work strengthens the connection between the machine learning and the neuroscience community. It~underlines the importance of predictive coding  in both areas, both as a highly plausible algorithm to better understand how memory and prediction work in the brain, and as an approach to solve corresponding problems in machine learning
and artificial intelligence. 


\section*{Acknowledgments}
This work was supported by the Alan Turing Institute under the EPSRC grant EP/N510129/1, 
by the AXA Research Fund, the EPSRC grant EP/R013667/1, and by the EU TAILOR grant. We also acknowledge the use of the EPSRC-funded Tier 2 facility
JADE (EP/P020275/1) and GPU computing support by Scan Computers International Ltd.

\bibliographystyle{ieeetr}
\bibliography{references}

\appendix

\newpage

\section{Detailed Derivation of Eqs.~\eqref{eq:pcn-dotx_il} and \eqref{eq:pcn-update-param}}

\subsection{\label{sec:il-inference}Derivation of Eq.~\eqref{eq:pcn-dotx_il}}

By expanding Eq.~\eqref{eq:pcn-f} with the definition of $\varepsilon^{\scriptscriptstyle {l}}_{i,t} = x^{\scriptscriptstyle {l}}_{i,t} - \mu^{\scriptscriptstyle {l}}_{i,t}$, we have:
\begin{align}
E_{t} = 
{\textstyle\sum}_{l=0}^{L-1} {\textstyle\sum}_{i=1}^{n^{\scriptscriptstyle {l}}}
{ \mbox{$\frac{1}{2}$} ( \varepsilon^{\scriptscriptstyle {l}}_{i,t} ) ^2}
= {\textstyle\sum}_{l=0}^{L-1} {\textstyle\sum}_{i=1}^{n^{\scriptscriptstyle {l}}}
{ \mbox{$\frac{1}{2}$} ( x^{\scriptscriptstyle {l}}_{i,t} - \mu^{\scriptscriptstyle {l}}_{i,t} ) ^2}\,.
\label{eq:pcn-f-expand}
\end{align}
Inference minimizes $E_{t}$ by modifying $x^{\scriptscriptstyle {l}}_{i,t}$ proportionally to
the gradient of the objective function~$E_{t}$.
We note that each $x^{\scriptscriptstyle {l}}_{i,t}$ influences $E_{t}$ in two ways: (i)~it occurs in Eq.~\eqref{eq:pcn-f-expand} explicitly, but (ii) it also determines the values of $\mu^{\scriptscriptstyle {l-1}}_{k,t}$ via Eq.~\eqref{eq:pcn-forward-varepsilon}. 
Therefore, 
the derivative of $E_{t}$ over $x^{\scriptscriptstyle {l}}_{i,t}$ contains two terms, which are formally as follows:
\begin{align}
\Delta{x}^{\scriptscriptstyle {l}}_{i,t} 
&= - \gamma\cdot\frac{\partial E_{t}}{\partial x^{\scriptscriptstyle {l}}_{i,t}} \\
&= - \gamma\cdot(\frac{\partial \mbox{$\frac{1}{2}$} ( x^{\scriptscriptstyle {l}}_{i,t} - \mu^{\scriptscriptstyle {l}}_{i,t} ) ^2}{\partial x^{\scriptscriptstyle {l}}_{i,t}} + \frac{\partial {\textstyle\sum}_{k=1}^{n^{\scriptscriptstyle {l-1}}} \mbox{$\frac{1}{2}$} ( x^{\scriptscriptstyle {l-1}}_{k,t} - \mu^{\scriptscriptstyle {l-1}}_{k,t} ) ^2}{\partial x^{\scriptscriptstyle {l}}_{i,t}}) \\
&= \gamma\cdot ( -(x^{\scriptscriptstyle {l}}_{i,t} - \mu^{\scriptscriptstyle {l}}_{i,t}) + f' ( x^{\scriptscriptstyle {l}}_{i,t} )  {\textstyle\sum}_{k=1}^{n^{\scriptscriptstyle {l-1}}} (x^{\scriptscriptstyle {l-1}}_{k,t} - \mu^{\scriptscriptstyle {l-1}}_{k,t}) \theta^{\scriptscriptstyle {l}}_{k,i} ) \\
&= \gamma\cdot ( -\varepsilon^{\scriptscriptstyle {l}}_{i,t} + f' ( x^{\scriptscriptstyle {l}}_{i,t} ) {\textstyle\sum}_{k=1}^{n^{\scriptscriptstyle {l-1}}} \varepsilon^{\scriptscriptstyle {l-1}}_{k,t} \theta^{\scriptscriptstyle {l}}_{k,i} )\,.
\end{align}
Considering also the special cases of $l=L$ and $l=0$, we obtain Eq.~\eqref{eq:pcn-dotx_il}.

\subsection{\label{sec:il-update}Derivations of Eq.~\eqref{eq:pcn-update-param}}

The update of weights minimizes $E_{t}$ by modifying $\theta^{\scriptscriptstyle {l+1}}_{i,j}$ proportionally to the gradient of the objective function $E_{t}$.
We note that $\theta^{\scriptscriptstyle {l+1}}_{i,j}$ affects the value of the function $E_{t}$ of Eq.~\eqref{eq:pcn-f-expand} by influencing $\mu^{\scriptscriptstyle {l}}_{i,t}$ via Eq.~\eqref{eq:pcn-forward-varepsilon}, hence,
the derivative of the objective function $E_{t}$ over $\theta^{\scriptscriptstyle {l+1}}_{i,j}$ can be formally defined as:
\begin{align}
\Delta \theta^{\scriptscriptstyle {l+1}}_{i,j} &= -\alpha\cdot {\partial E_{t}}/{\partial \theta^{\scriptscriptstyle {l+1}}_{i,j}} \\
&= -\alpha\cdot \frac{\partial \mbox{$\frac{1}{2}$} ( x^{\scriptscriptstyle {l}}_{i,t} - \mu^{\scriptscriptstyle {l}}_{i,t} ) ^2}{\partial \theta^{\scriptscriptstyle {l+1}}_{i,j}}  \\
&=  \alpha\cdot \varepsilon^{\scriptscriptstyle {l}}_{i,t} f ( x^{\scriptscriptstyle {l+1}}_{j,t} )\,.
\end{align}

\section{Further Details on the Experiments}

Here, we provide further details about training PCNs, useful to reproduce them. Particularly, every PCN was trained using IL until convergence, we varied the following hyperparameters and reported the best result: $T \in \{12,16,24,32\}$ for $2$-layer networks, and $T \in \{32,48,72\}$ for multilayer ones, $\gamma \in \{1,0.5,0.1,0.05,0.01\}$, and $\alpha \in \{0.0001,0.00005\}$. 

Furthermore, we used standard PyTorch initialization and ReLU activations in every layer. To retrieve corrupted images, we have iterated the function $F$ $30$ times, using $T \in \{ 100,250,500\}$, and reported the best results.

\newpage

\section{Analysis of Different Levels of Noise}

\begin{figure}[H]
\includegraphics[width=0.98\linewidth]{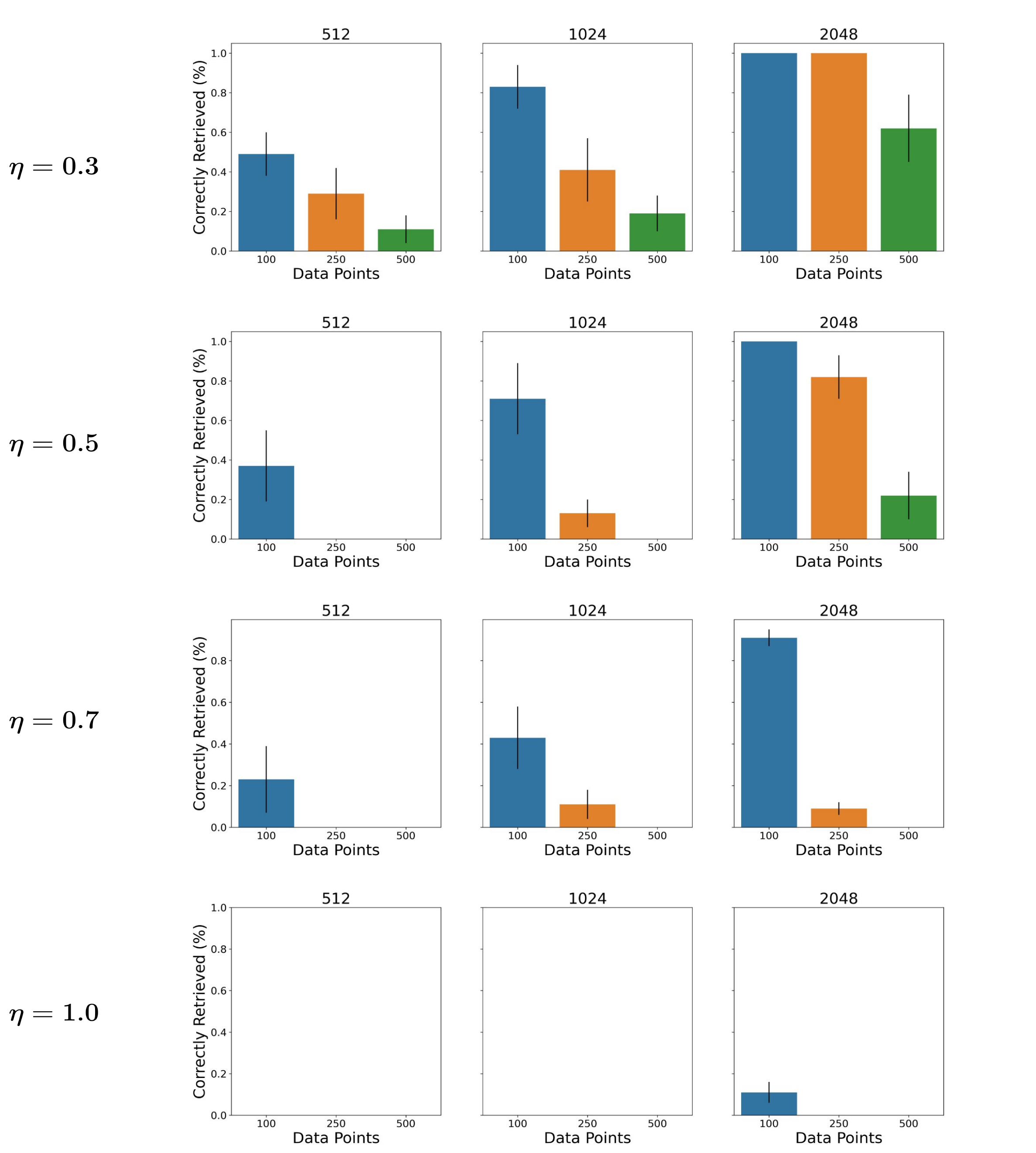}
\caption[short]{Percentage of correctly retrieved images by 2-layer generative PCNs (PC) with hidden-layer dimensions of 512, 1024, and 2048, when presented with a corrupted image with Gaussian noise of different variances, trained on the Tiny ImageNet dataset.}
\label{fig:gauss_supp}\vspace*{-1ex}
\end{figure}

So far, we analyzed images with Gaussian noise of variance $0.2$. We now extend the analysis to different levels of corruptions. Particularly, we have used networks of width $n \in \{512,1024,2048\}$ trained on $N \in \{100,250,500\}$ natural images taken from the Tiny ImageNet dataset, and tried to retrieve them using Gaussian noise of variance $\eta \in \{ 0.3,0.5,0.7,1.0\}$. 

\textbf{Results:} The results show that our method is able to retrieve stored data points even with a larger amount of variance, although the results get worse the more we increase the level of noise. Particularly, when presented with data points with Gaussian noise with variance $0.3$, every network was able to restore at least one image, even when trained on $500$ examples. The numbers of retrieved images decreased when increasing the noise. When presented with noise with variance $\eta = 1.0$, only networks with $2048$ hidden units were able to retrieve a tiny fraction of the original data points. We report the results in Fig.~\ref{fig:gauss_supp}

\section{Plots of Images with Extreme Levels of Noise}

\begin{figure}[H]
\centering
\includegraphics[width=0.85\textwidth]{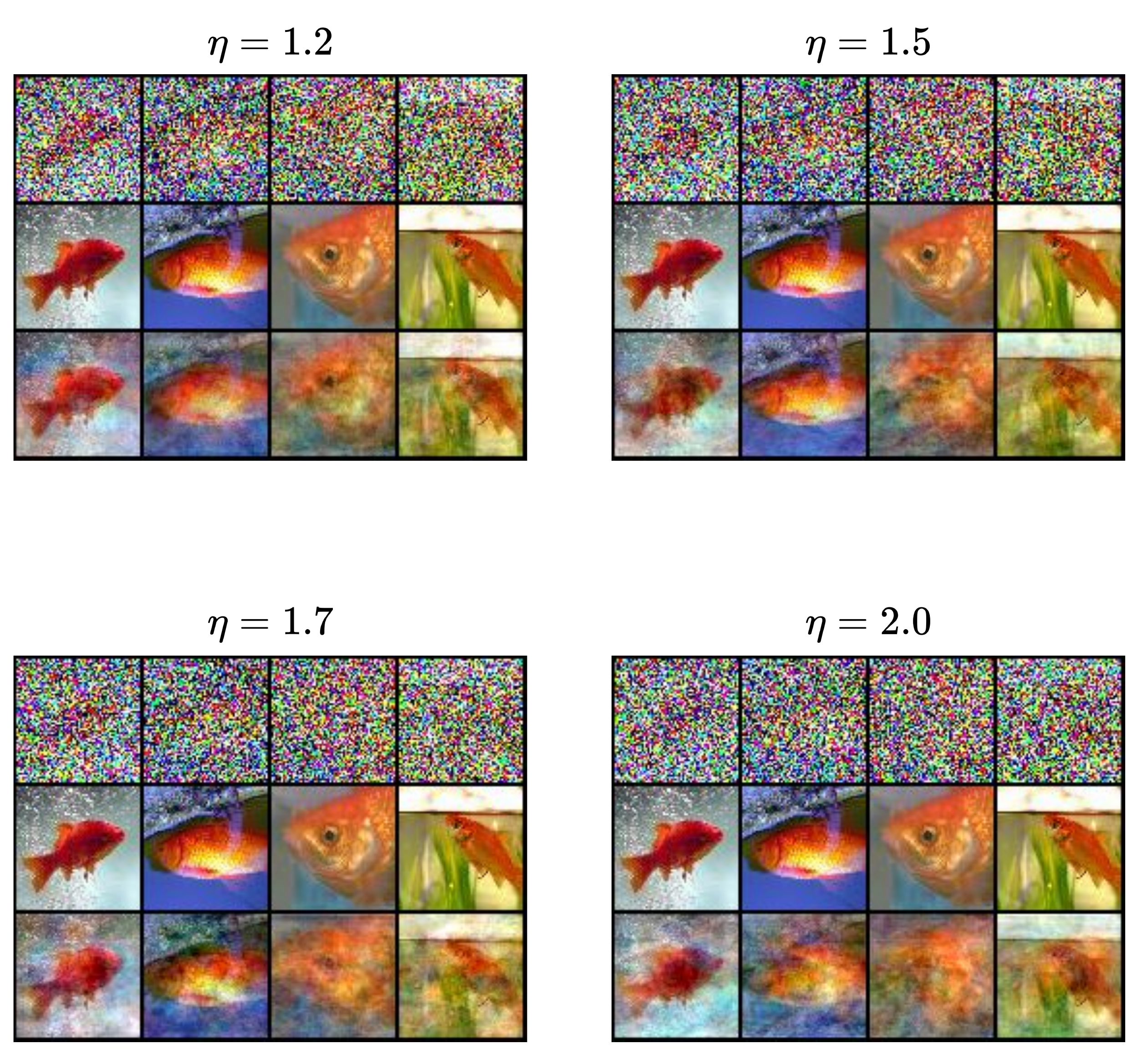}
\caption[short]{Analysis of the reconstruction capabilities of generative PCNs under different levels of noise $\eta \in \{1.2,1.5,1.7,2.0\}$. Every plot was obtained using a $2$-layer generative PCN with $2048$ hidden units, trained on $100$ images of the Tiny ImageNet dataset.}
\label{fig:extreme}
\end{figure}

We now plot the images generated using different levels of corruptions. Note that all the images shown here were clearly classified as \emph{not} retrieved by the above analysis. However, we believe that an analysis of not well-retrieved images is still interesting, to understand the limitations of our method. Hence, we  trained a generative PCN on $100$ images of the Tiny ImageNet dataset, and tried to reconstruct them using extreme levels of noise. Then, we printed the reconstructions.

\textbf{Results:} These representations show that generative PCNs are able to identify the original data points, even if the retrieval process was not accurate enough for the error to fall below the decided threshold $0.005$. Particularly, our model was able to identify original memories even when presented with extreme levels of noise $(\eta \in \{1.2,1.5,1.7,2.0\})$, although leaving visible amount of corruption. The reconstructions are given in Fig.~\ref{fig:extreme}.

\section{Analysis of the Retrieval Function}

\begin{figure}
\includegraphics[width=1\textwidth]{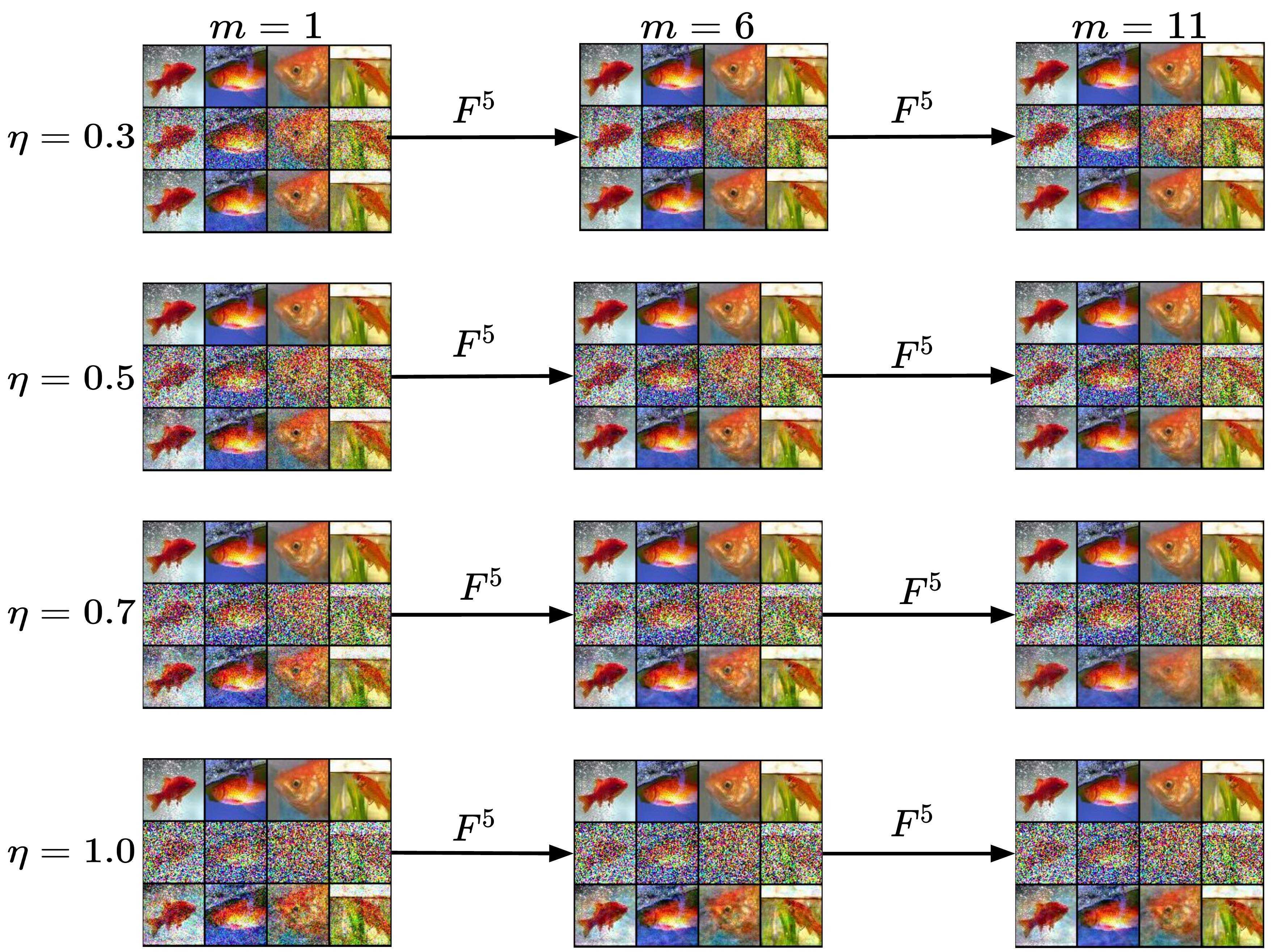}
\caption[short]{Analysis of the reconstruction capabilities of $10$ iterations of the function $F$ under different levels of noise $\eta \in \{0.3,0.5,0.7,1.0\}$. Particularly, we  plotted the results after $1$, $6$,  and $11$ iterations of $F$.}
\label{fig:F}\vspace*{-1ex}
\end{figure}

Here, we show visually how the function $F$ restores corrupted images. Particularly, we  trained a generative PCN with $2048$ hidden neurons on $100$ images of the Tiny ImageNet dataset, and printed the reconstructions after $1$, $6$, and $11$ iterations of the retrieval function $F$. 

\textbf{Results:} The reconstructions show that the first iteration is able to clear most of the noise. However, further iterations are needed to clear the remaining details, especially if the noise level is high. In fact, it can be observed that one iteration of $F$ is able to retrieve the original data point when the level of corruption is $\eta = 0.3$, but fails when $\eta=1.0$. This process usually converges around $m = 15$, depending on the number of iterations $T$ and the level of noise. The results are shown in Fig.~\ref{fig:F}.

\newpage

\section{Reconstruction of Partial Images}

\begin{figure}[h]
\centering
\includegraphics[width=1\textwidth]{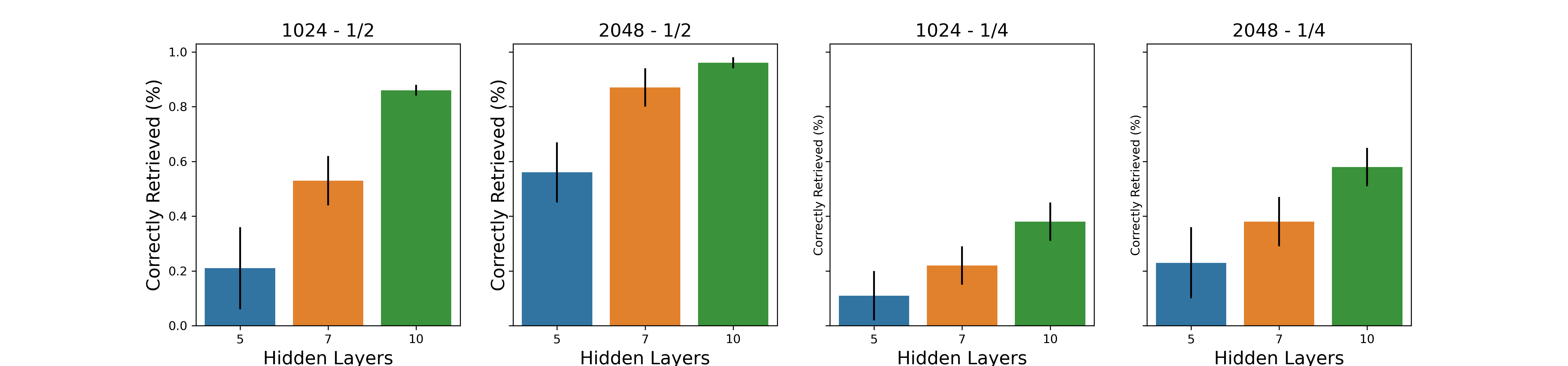}

\caption[short]{
Percentage of correctly retrieved images as a function of the number of hidden layers, for generative PCNs on a dataset of $500$ images of the CIFAR10 dataset. We used networks with $n=\{1024,2048\}$ hidden neurons, and tried to reconstruct them using partial images containing a fraction of $1/2$ and $1/4$ of the original pixels.}
\label{fig:depth}\vspace*{-2ex}
\end{figure}

To further show the robustness of our experiments, we replicate the experiments of Section 5.1 on the CIFAR10 dataset, training networks of different depths ($L\in \{5,7,10\})$ to generate $500$ images, and report the results in Fig.~\ref{fig:depth}. Again, we consider a data point to be perfectly reconstructed if the error between the original image and the reconstruction is below $0.001$. 

\textbf{Results:} The plots confirm  the results stated in the body of this work: the deeper the network, the more images are perfectly reconstructed.

\section{Analysis of the Retrieval Time}

\begin{table}[h]
\caption{Time (in seconds) to perform a retrieval from corrupted images with injected Gaussian noise of variance $\eta = 0.2$.}
  \label{tb:div}\vspace*{1ex}
  \centering
\resizebox{0.5\columnwidth}{!}{
    \begin{tabular}{ccccc}
    \toprule
     Hidden Dimension & $256$ & $512$ & $1024$  & $2048$     \\
    \midrule
    One Iteration    & $0.02$   & $0.04$ & $0.09$ & $0.15$   \\
    Five Iterations  & $0.10$   & $0.19$ & $0.53$ & $0.84$   \\
    Ten Iterations   & $0.55$   & $1.18$ & $2.85$ & $4.37$   \\
    \bottomrule
    \end{tabular}
    \label{tab:time_corr}
  }
\end{table}

\begin{table}[h]
\caption{Time (in seconds) to perform a retrieval from an incomplete image, where only $1/2$ of the original pixels are provided.}
  \label{tb:div}\vspace*{1ex}
  \centering
\resizebox{0.5\columnwidth}{!}{
    \begin{tabular}{ccccc}
    \toprule
     Hidden Dimension & $256$ & $512$ & $1024$  & $2048$     \\
    \midrule
    Error = $0.005$    &  $0.74$  & $1.08$ & $1.83$  & $2.48$  \\
    Error = $0.001$  & $14.69$   & $24.19$ & $37.73$ & $53.12$   \\
    \bottomrule
    \end{tabular}
    \label{tab:time_par}
  }
\end{table}

In the AM literature, it is important to quickly retrieve an image. In fact, modern Hopfield networks and similar models are able to retrieve memories 1-shot. Our model is slower than other AM models, as the retrieval process relies on an energy minimization framework, however, it is able to retrieve a memory in a couple of seconds. Note that this does not alter its biological plausibility, mainly because of two reason:

\begin{itemize}
    \item The brain performs computations much more efficiently, and hence the same process could take milliseconds. The neural information processing is, in fact, certainly not based on digital GPU/CPU hardware and may realize the presented algorithm in a much more natural (and efficient) way, such as using analogue relaxation dynamics to rapidly converge to the attractor state.
    
    \item The retrieval process takes some time, because it retrieves exact memories at the pixel level, while memory recall is not as detailed. If the inference process is stopped after a shorter period of time, it simply recalls a slightly fuzzier memory. This flexibility of variable-length computation is also highly desirable and realistic from a biological viewpoint.
    
\end{itemize}

The time in seconds needed to retrieve original images from corrupted data is given in Table~\ref{tab:time_corr}, while the time needed to retrieve original images from incomplete data is given in Table~\ref{tab:time_par}. Note that these numbers provide an upper bound on the real efficiency of the model, limited by the current implementation. In fact, while every computation during the retrieval process can theoretically run in parallel, this cannot be implemented in popular deep learning frameworks such as PyTorch and Tensorflow. A correct implementation, which would require significant engineering work, going beyond the scope of this paper, could make this process  significantly faster.

All experiments are conducted on two Nvidia GeForce GTX 1080Ti GPUs and eight Intel Core i7 CPUs, with 32 GB RAM.

\section{Full-Page Reconstructions of ImageNet}

To show the reconstruction capabilities of our method, in what follows we plot full-page reconstructions of ImageNet pictures.

\begin{figure}[t!]
\includegraphics[width=1.0\linewidth]{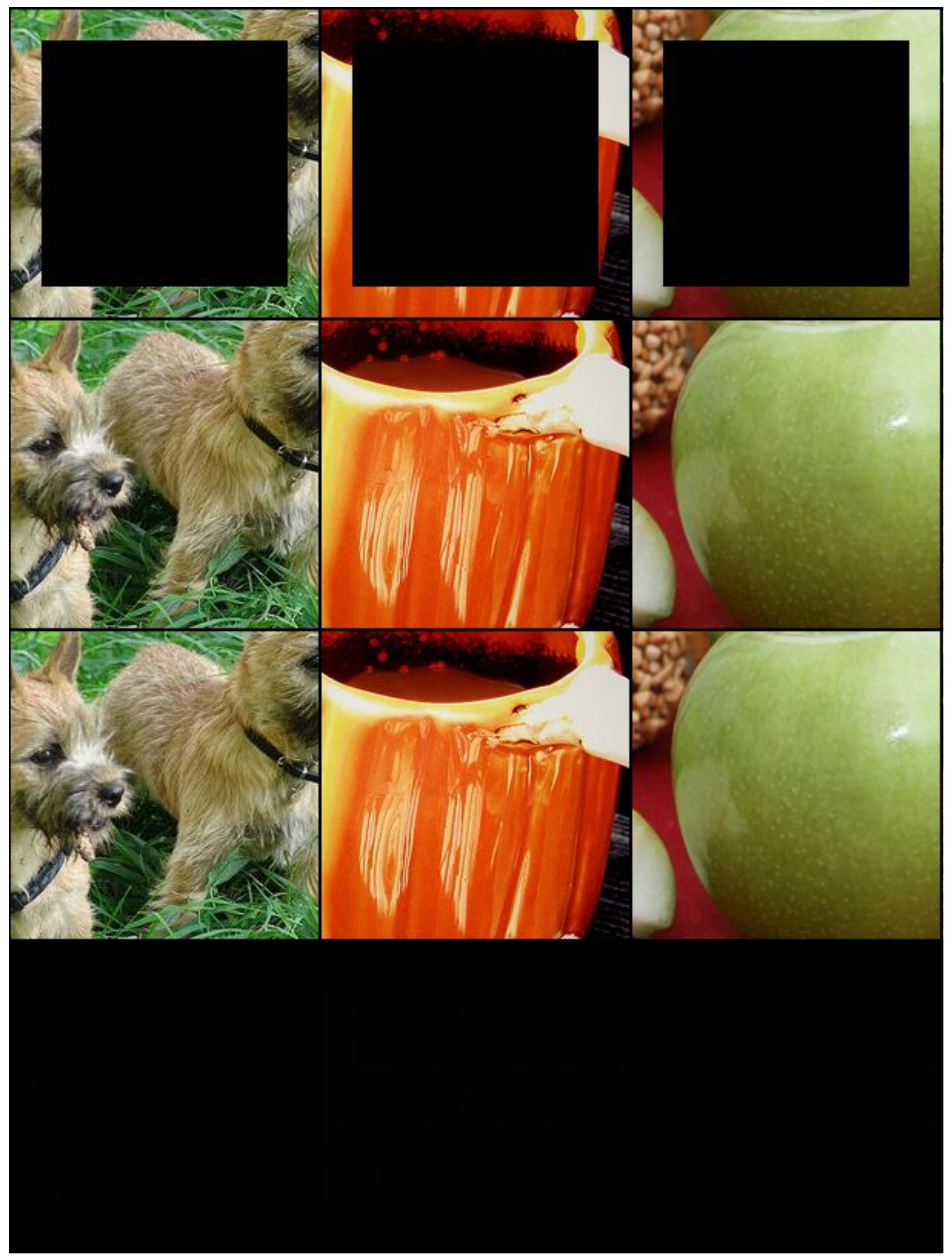}
\caption[short]{To show the level of accuracy of our proposed method, we represent reconstructions of ImageNet pictures. Here, a reconstruction when a patched image is presented to our generative PCN. 
From top to bottom: partial image, reconstructed
image, original image, and reconstruction error (difference between original and reconstruction).}
\label{fig:IMG1}\vspace*{-1ex}
\end{figure}

\begin{figure}[t!]
\includegraphics[width=1\textwidth]{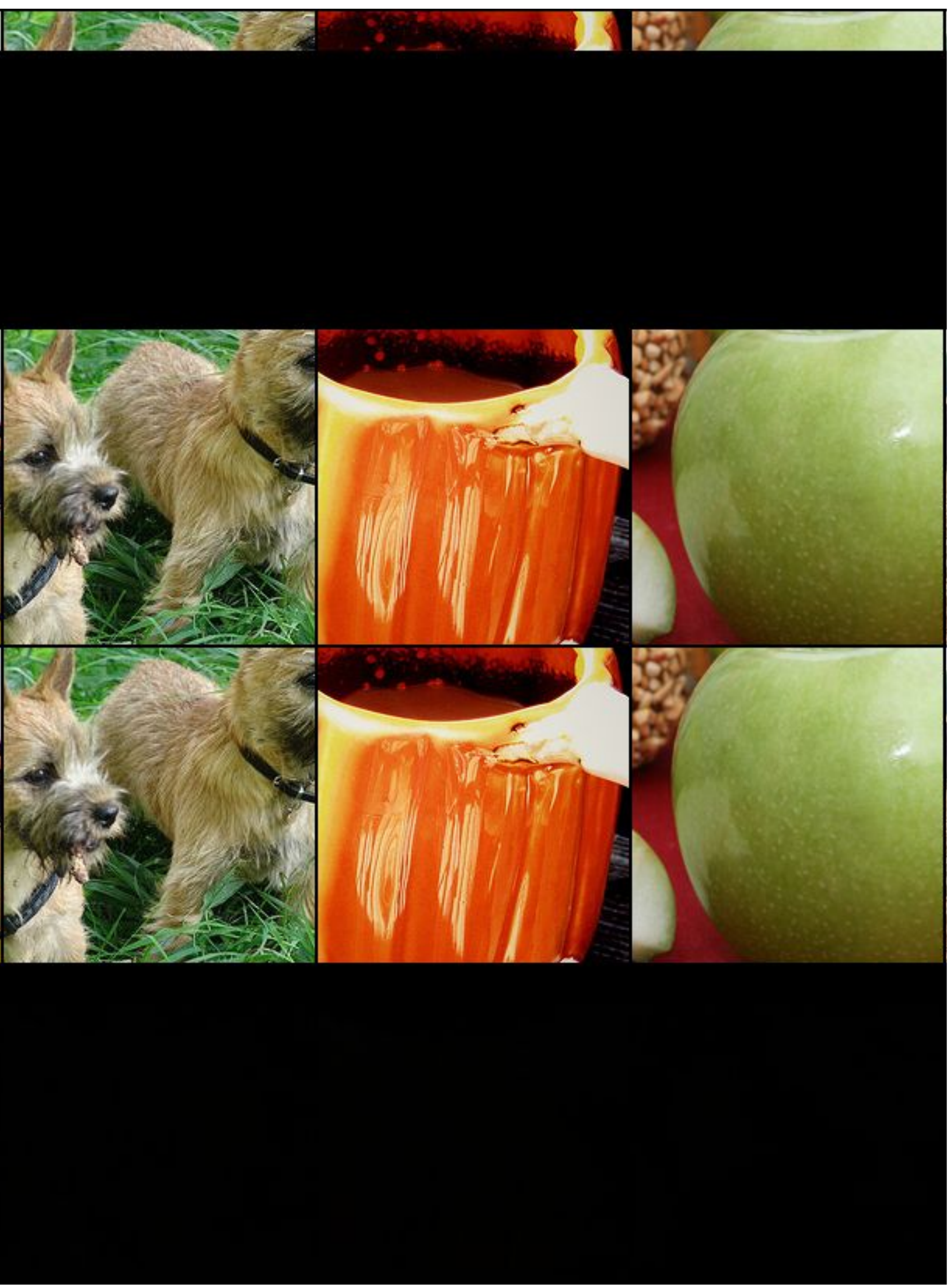}
\caption[short]{To show the level of accuracy of our proposed method, we represent reconstructions of ImageNet pictures. Here, a reconstruction when only $1/8$ of the original picture is presented to our generative PCN.
From top to bottom: partial image, reconstructed
image, original image, and reconstruction error (difference between original and reconstruction).}
\label{fig:IMG2}\vspace*{-1ex}
\end{figure}


\end{document}